\documentclass[11pt,letterpaper]{article}

\usepackage[margin=1in]{geometry}
\usepackage[T1]{fontenc}
\usepackage{microtype}

\usepackage{amsmath,amssymb,amsfonts}
\usepackage{algorithm}
\usepackage{algorithmic}


\usepackage{graphicx}
\usepackage{subfig}
\usepackage{booktabs,makecell,multirow,tabularx,array}
\usepackage{caption}
\usepackage{adjustbox}

\usepackage[numbers,sort&compress]{natbib}
\usepackage{url}
\usepackage{xcolor}
\usepackage[colorlinks=true,linkcolor=blue,citecolor=blue,urlcolor=blue]{hyperref}
\hypersetup{
  pdftitle={Learning Interpretable Differentiable Logic Networks for Tabular Regression},
  pdfauthor={Chang Yue and Niraj K. Jha},
  pdfsubject={Interpretable machine learning for tabular regression},
  pdfkeywords={edge computing, interpretability, logic rules, machine learning, neural networks, regression},
  pdfcreator={LaTeX}
}

\providecommand{\Description}[1]{}

\newcolumntype{Y}{>{\centering\arraybackslash}X}
\newcolumntype{C}[1]{>{\centering\arraybackslash}p{#1}}
\newcolumntype{L}{>{\raggedright\arraybackslash}X}

\newcommand{\meanstd}[2]{%
  \begin{tabular}{@{}c@{}}
    #1 \\[-1pt]
    \scriptsize$\pm$\,#2
  \end{tabular}%
}

\title{Learning Interpretable Differentiable Logic Networks for Tabular Regression}
\author{%
  Chang Yue \qquad Niraj K. Jha\\[0.4em]
  \small Princeton University, Princeton, NJ, USA\\[-0.1em]
  \small \texttt{cyue@princeton.edu} \qquad \texttt{jha@princeton.edu}
}
\date{}

\begin{document}
\maketitle

\begin{abstract}
Neural networks (NNs) achieve outstanding performance in many domains; however, their decision processes are often
opaque and their inference can be computationally expensive in resource-constrained environments. We recently proposed
Differentiable Logic Networks (DLNs) to address these issues for tabular classification based on relaxing discrete logic
into a differentiable form, thereby enabling gradient-based learning of networks built from binary logic operations.
DLNs offer interpretable reasoning and substantially lower inference cost.

We extend the DLN framework to supervised tabular regression. We first redesign the final output layer (the SumLayer)
to support continuous targets. More critically, we find the original two-phase training procedure used for
classification is suboptimal for regression, and thus develop a unified, single-stage optimization procedure.
We also demonstrate that temperature annealing of the network's differentiable relaxations is
decisive for achieving stable convergence and high accuracy. We evaluate the resulting model on 15 public
regression benchmarks, comparing it with modern neural networks and classical regression baselines. Regression DLNs
match or exceed baseline accuracy while preserving interpretability and fast inference. Our results show that
DLNs are a viable, cost-effective alternative for regression tasks, especially where model transparency and
computational efficiency are important.
\end{abstract}

\vspace{0.6em}

\section{Introduction}
Interpretable neural architectures that integrate logical reasoning have gained increasing attention in 
recent years. Petersen \emph{et al.}~\cite{petersen2022deep} first proposed \emph{Deep Differentiable Logic Gate Networks}
(LGNs), which relax the learning of conventional Boolean logic operations (e.g., AND, XOR) into differentiable 
forms, thereby enabling gradient-based learning of networks composed of logic gates. This strategy yields fast, 
interpretable models: the logic structure permits extraction of human-readable rules and the discrete implementation delivers 
highly efficient inference.

Building on this foundation, Petersen \emph{et al.}~\cite{petersen2024convolutional} introduced
\emph{Convolutional Differentiable Logic Gate Networks}, adding deep logic-gate tree convolutions, logical-OR
pooling, and residual initializations. These enhancements enable LGNs to exploit the convolution paradigm and scale
to more complex classification tasks with improved accuracy. For example, their convolutional LGNs achieve
86.29\% accuracy on CIFAR-10 with a model that is 29$\times$ smaller than the previous state of the art.

Yue and Jha~\cite{10681646} restructured the LGN framework for general tabular classification, eliminating
various constraints, such as fixed network topologies and binary-only inputs. The resulting \emph{Differentiable Logic
Networks} (DLNs) match or surpass standard NNs and LGNs on 20 public benchmarks while retaining
interpretability and operating at several orders of magnitude lower computational cost than conventional multilayer
perceptrons (MLPs).

Collectively, these studies establish logic-gate networks as a compelling alternative to conventional neural
architectures in classification settings, combining transparency with efficiency.

Despite these advances, existing logic-gate networks are limited to \emph{classification}, predicting discrete class
labels. Yet, many real-world applications in the scientific and industrial domains require \emph{regression}, that is,
predicting a continuous numerical output from input features. Current DLN architectures cannot be applied directly to
such problems because they generate class scores or probabilities rather than a single continuous value. For instance,
a DLN classifier counts the number of learned logical rules that fire for each class and selects the class with the
most activated rules. Consequently, there is strong motivation to extend the DLN paradigm beyond classification,
enabling tabular regression while preserving the interpretability and speed advantages that distinguish DLNs.

In this article, we introduce a regression-focused extension of DLNs. The core innovation is a new \textit{SumLayer}
that enables DLNs to output a continuous value from the final layer of logic neurons. Specifically, the binary outputs
of the last hidden layer are linearly combined by a SumLayer that learns continuous weights, connecting those outputs
to a single numerical node. Unlike prior DLN classifiers, which aggregate logic activations into discrete class counts
or probabilities, our SumLayer computes a weighted sum of the activated rules, yielding a continuous prediction. This
simple, yet effective, change enables DLNs to address regression tasks directly. Crucially, the model’s interpretability
is preserved: each neuron in the last hidden layer still represents a logical rule learned through differentiable
relaxation and the final output is an intelligible weighted sum of these binary evaluations.

\begin{figure}[!tp]
\centering
\includegraphics[width=0.8\textwidth]{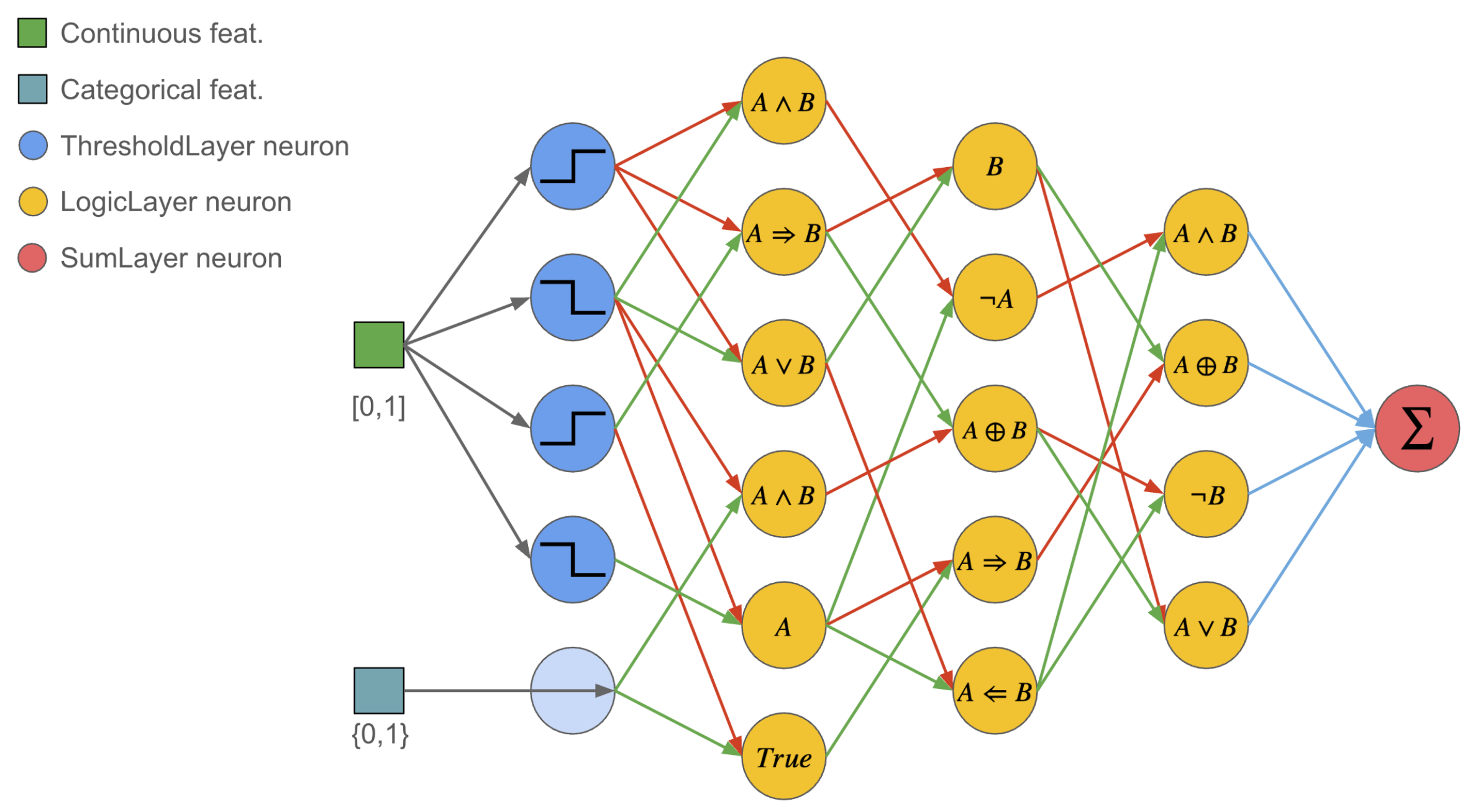}
\caption{A simplified regression DLN example. Continuous input features are first binarized by a
ThresholdLayer, producing a binary vector. This vector is then processed by successive LogicLayers composed of
two-input Boolean operators. The activations of the final LogicLayer are combined by a SumLayer that computes a
weighted sum, yielding the real-valued prediction.}
\label{example}
\Description{A flowchart illustrating the architecture of a regression DLN. 
It shows continuous inputs being binarized, processed through logic layers 
with logic gates, and finally combined by a sum layer to produce a continuous output.}
\end{figure}

Beyond this architectural change, we also enhance the model's optimization. While prior
classification DLNs rely on a two-phase training schedule~\cite{10681646}, we find this
approach to be suboptimal for the continuous optimization required in regression. We, therefore,
employ a unified, single-stage optimization procedure.
This new training methodology, combined with a temperature annealing schedule that our
ablation studies show is decisive for stable convergence,
enables end-to-end learning of logical rules for regression tasks. Furthermore, we extend
the original DLN work by evaluating our model's performance under ensembling and
providing a rigorous inference latency analysis using Ahead-of-Time (AOT) compiled
code, confirming its practical efficiency. A simplified regression DLN is shown in Fig.~\ref{example}.

We empirically validate the proposed model on 15 benchmark tabular regression datasets. Across these tasks, the
regression DLN achieves predictive accuracy comparable to state-of-the-art black-box models, despite its much smaller
logical architecture.

The remainder of this article is organized as follows. Section~\ref{sec-related} reviews related work.
Section~\ref{sec-method} describes the regression DLN architecture and the SumLayer mechanism.
Section~\ref{sec-experiments} presents experimental results and analysis. Section~\ref{sec-conclusion} concludes
with a summary of findings and directions for future research.

\section{Related Work}\label{sec-related}
This work on regression DLNs intersects with several active research areas in machine learning: differentiable
models, interpretable AI systems, and efficient machine learning.

\textbf{Differentiable Machine Learning:} The core idea of making traditionally non-differentiable structures
differentiable is to enable gradient-based optimization. Real-valued logic extends classical Boolean logic to operate
on continuous values (typically between 0 and~1) rather than on discrete true/false statements~\cite{zadeh1978fuzzy,menger2003statistical,goertzel2008probabilistic}. Efforts to create differentiable decision
trees~\cite{frosst2017distilling,yang2018deep} or embed them within larger differentiable
architectures~\cite{kontschieder2015deep,tanno2019adaptive,popov2019neural} aim to pair the interpretability of tree
structures with the representational power of deep learning. Differentiable modeling is also being explored as a
bridge between machine learning and scientific computing, allowing domain knowledge and constraints to be integrated
directly into the learning process~\cite{liu2024kan,raissi2019physics,hu2019difftaichi,cranmer2020discovering,baydin2018automatic}. 
Continuous relaxation techniques are pivotal in this context, transforming discrete operations into smooth, differentiable
counterparts so that gradients can be computed reliably~\cite{petersen2022learning,cuturi2017soft}. Building on these principles, 
logic-gate-based neural networks have made rapid progress~\cite{petersen2022deep,petersen2024convolutional,10681646}. Our work 
extends this line of research to tabular regression tasks by introducing a differentiable logic-gate architecture capable of 
predicting continuous targets.

\textbf{Interpretable Machine Learning:} The quest for transparency has spurred extensive work on explaining the
behaviour of black-box models. Popular post-hoc techniques include Local Interpretable Model-Agnostic 
Explanations~\cite{ribeiro2016should}, Shapley Additive
Explanations~\cite{lundberg2017unified}, saliency maps~\cite{simonyan2013deep}, feature
visualization~\cite{olah2017feature}, and, for regression, partial dependence plots and individual conditional
expectation plots~\cite{goldstein2015peeking}. Post-hoc explanations, however, can be unreliable or even
misleading~\cite{rudin2019stop}, motivating the design of inherently interpretable models.  Classical examples
for regression include linear models and their regularized variants, Ridge and
Lasso~\cite{hoerl1970ridge,tibshirani1996regression}, whose coefficients offer direct insight.  More expressive,
yet still transparent, approaches comprise Generalised Additive Models~\cite{hastie2017generalized}, which capture nonlinear 
effects additively, and rule-based systems such as
RuleFit~\cite{friedman2008predictive}.  Broader interpretable structures include sparse decision
trees~\cite{lin2020generalized} and scoring systems~\cite{ustun2016supersparse}.
Neurosymbolic methods seek to fuse symbolic reasoning with neural representations. Logical Neural Networks~\cite{riegel2020logical} 
construct differentiable logic circuits with learnable weights but require a
pre-specified logical skeleton, which may demand considerable domain expertise.  Other efforts learn logic rules
directly: Wang \emph{et al.}~\cite{wang2021scalable} propose the Rule-based Representation Learner, projecting rules
into a continuous space and introducing Gradient Grafting for differentiation.  Neural Logic
Machines~\cite{dong2019neural} perform multi-hop inference with iterative layers that approximate logical
quantifiers, while Neural Logic Networks~\cite{shi2019neural} dynamically build neural computation graphs from
logical expressions and learn basic operations for propositional reasoning.
Our regression DLN follows the same principle of end-to-end differentiability as the classification DLN but adapts
the architecture to continuous targets.  By learning weighted contributions from the final logic layer through a
SumLayer, the model provides a transparent mapping from binary rule activations to the predicted value.  The entire
network remains optimizable with gradient descent while offering interpretability that is often missing from complex
regression models.

\textbf{Efficient Machine Learning:}  The aim of efficient machine learning is to reduce the computational and memory
footprints of modern models, easing deployment on resource-constrained hardware, and lowering operational costs.  
While our model's training process is computationally intensive, its final architecture is designed
for high inference efficiency, which is a critical requirement for many applications. 
Key
strategies include network pruning, which removes redundant parameters or connections, to leave a sparse,
lightweight subnetwork~\cite{han2015learning,frankle2018lottery}. Quantization further shrinks models and can
speed up inference by lowering the numerical precision of weights and activations. Extreme variants enable binary
representations, as in Binarized Neural Networks (BNNs)~\cite{courbariaux2016binarized} and
XNOR-Net~\cite{rastegari2016xnor}, while recent work focuses on large-model quantization techniques, such as
Generative Pre-trained Transformer Quantization~\cite{frantar2022gptq,zhao2024atom}. Knowledge distillation offers 
another route to efficiency: A small
student network learns to mimic the outputs and internal representations of a larger teacher, yielding
compact models that retain much of the teacher's accuracy~\cite{hinton2015distilling,gu2023minillm}. 
While these methods produce smaller and faster systems, their goals and domains differ from ours.
BNN and XNOR-Net, for example, are foundational techniques for compressing existing black-box
models, like convolutional neural networks, and are overwhelmingly applied to large-scale computer vision tasks. Although these
methods produce smaller and faster systems, they do not automatically enhance interpretability.
Hardware specialization provides another avenue for speed and energy savings. Application-Specific Integrated
Circuits (ASICs) and Field-Programmable Gate Arrays (FPGAs) deliver task-specific acceleration; for instance, Umuroglu
\emph{et al.}~\cite{umuroglu2017finn} and Petersen \emph{et al.}~\cite{petersen2024convolutional} report substantial
speed-ups for NNs on FPGAs. Because our models rely on simple logic operations and sparse connectivity,
they are promising candidates for hardware acceleration on these platforms, offering a path to high efficiency 
without sacrificing interpretability.

\section{Methodology}\label{sec-method}
This section first offers a high-level overview of our approach and then details how each layer is trained.  Four
changes distinguish the proposed regression DLN from its classification predecessor.  First, the SumLayer
produces a floating-point weighted sum of logic activations rather than a discrete class-wise vote.  Second, the
network ends in a single-output neuron that outputs a real-valued prediction instead of one node per class.  Third,
neuron functions and connections are learned jointly in a single optimization stage, replacing the two-phase
procedure used for classification DLNs.  Fourth, every sigmoid and softmax employs a temperature
parameter that is annealed during training, gradually sharpening the approximations to hard logical decisions.

\begin{figure}[!htbp]
\centering
\includegraphics[width=0.3\columnwidth]{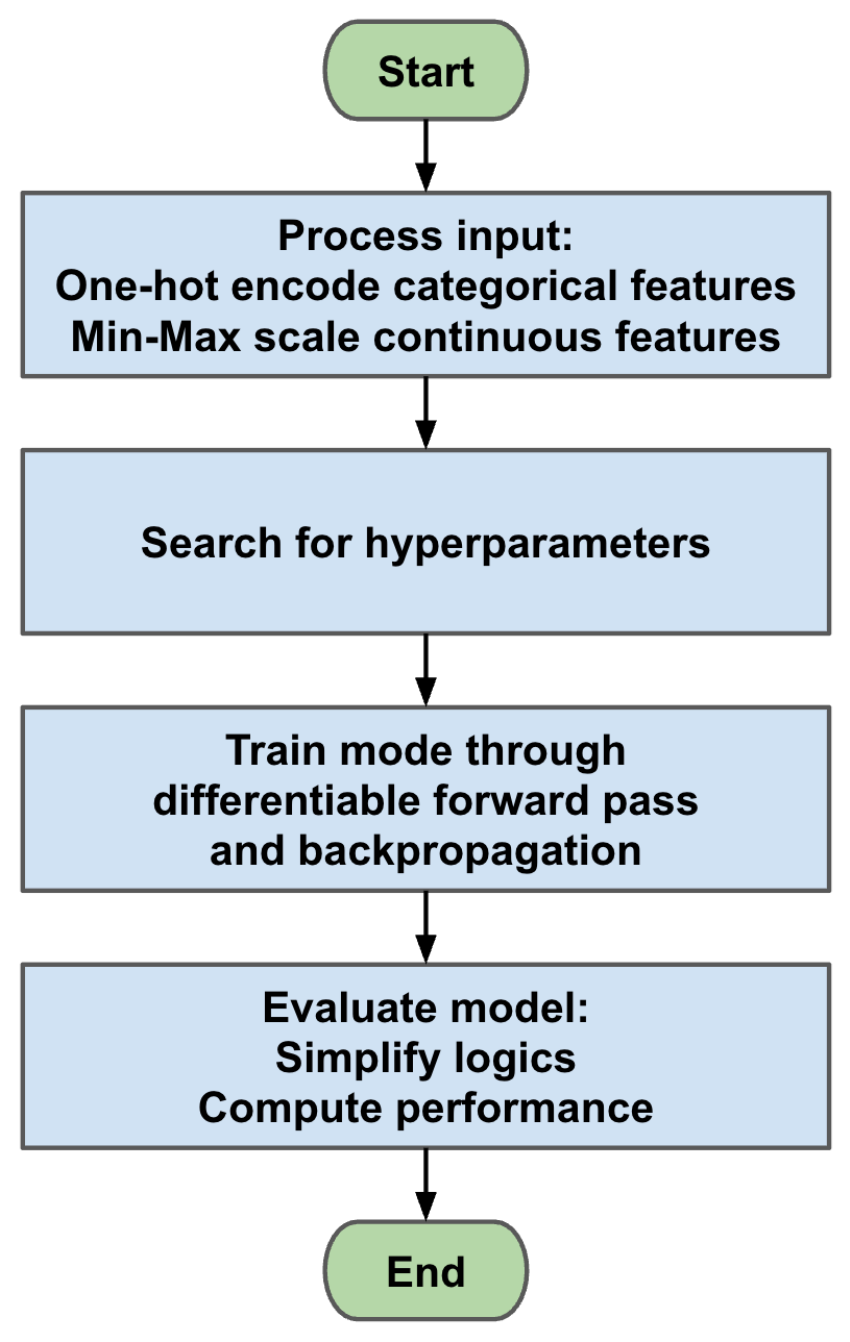}
\caption{Training workflow for a regression DLN.}
\label{flowchart}
\Description{A flowchart diagram outlining the steps involved in training a regression DLN.}
\end{figure}

Fig.~\ref{flowchart} summarises the complete training workflow.  First, we pre-process and discretize raw inputs; we then
perform hyperparameter optimization (HPO) under a fixed computational budget.  Using the best hyperparameters, we train the
model end-to-end with gradient descent.  Finally, we simplify the learned logic expressions and evaluate the resulting predictor.

Similar to classification DLNs, the regression DLN, as shown in Fig.~\ref{example}, contains three layer types.
A \textit{ThresholdLayer} connected to the inputs binarizes continuous features into $0/1$ values.  One or more
\textit{LogicLayers} introduce nonlinearity through two-input Boolean operators. A final \textit{SumLayer}
aggregates the weighted activations of the last LogicLayer to produce a continuous output. 
This structure is inherently interpretable and, due to its sparsity and reliance on binary logic, 
results in highly efficient inference.

\newcommand{\cellcontent}[1]{%
  \begin{minipage}{\linewidth}
    \raggedright
    #1
  \end{minipage}}
\newcommand{\phspace}{\hspace{1em}}

\begin{table}[t!]
    \centering
    \caption{Summary of DLN trainable parameters and feedforward functions during unified training and during inference. We use \( \mathbf{x} \) to denote input and \( \mathbf{y} \) to denote the output of each layer.}
    \label{layer-params}
    \footnotesize{
    \begin{tabular}{>{\centering\arraybackslash}p{1.7cm} >{\raggedright\arraybackslash}p{3.4cm} >{\raggedright\arraybackslash}p{4.7cm} >{\raggedright\arraybackslash}p{5.2cm}}
        \toprule
          & ThresholdLayer & LogicLayer & SumLayer \\
        \midrule
        \cellcontent{Trainable\\parameters} & 
        \cellcontent{bias \( \mathbf{b} \in \mathbb{R}^{\text{in\_dim}} \) \\ slope \( \mathbf{s} \in \mathbb{R}^{\text{in\_dim}} \)} & 
        \cellcontent{logic fn weight \( \mathbf{W} \in \mathbb{R}^{\text{out\_dim} \times 16} \) \\ link \( a \) weight \( \mathbf{U} \in \mathbb{R}^{\text{out\_dim} \times \text{in\_dim}} \) \\ link \( b \) weight \( \mathbf{V} \in \mathbb{R}^{\text{out\_dim} \times \text{in\_dim}} \)} & 
        \cellcontent{link weight \( \mathbf{S} \in \mathbb{R}^{\text{in\_dim}} \) \\ coefficient \( \mathbf{C} \in \mathbb{R}^{\text{in\_dim}} \) } \\
        \midrule
        \cellcontent{Training} & 
        \cellcontent{\( \mathbf{y}_i = \mathrm{Sigmoid}\bigl(\mathbf{s}_i \cdot (\mathbf{x}_i - \mathbf{b}_i)/\tau\bigr) \)} & 
        \cellcontent{\( \mathbf{y}_i = \sum_{k=0}^{15} P_k \,\mathrm{SoftLogic}_k(a, b) \) \\ 
        \( P = \mathrm{Softmax}(\mathbf{W}_{i,:} / \tau) \) \\[3pt]
        \( a = \sum_{j=0}^{\text{in\_dim}-1} [\mathrm{Softmax}(\mathbf{U}_{i,:} / \tau)]_j \cdot \mathbf{x}_j \) \\ 
        \( b = \sum_{j=0}^{\text{in\_dim}-1} [\mathrm{Softmax}(\mathbf{V}_{i,:} / \tau)]_j \cdot \mathbf{x}_j \)} & 
        \cellcontent{\( y = \sum_{j=0}^{\text{in\_dim}-1} \mathrm{Sigmoid}\bigl(\mathbf{S}_{j}/\tau\bigr) \cdot \mathbf{C}_{j} \cdot \mathbf{x}_j \)} \\
        \midrule
        \cellcontent{Inference} & 
        \cellcontent{\( \mathbf{y}_i = \mathrm{Heaviside}\bigl(\mathbf{s}_i \cdot (\mathbf{x}_i - \mathbf{b}_i)\bigr) \)} & 
        \cellcontent{\( \mathbf{y}_i = \text{HardLogic}_k (\mathbf{x}_{a_i}, \mathbf{x}_{b_i}) \)\\
        \( k = \text{arg max}_{j} \mathbf{W}_{i,j} \)\\
        \( a_i = \text{arg max}_{j} \mathbf{U}_{i,j} \)\\
        \( b_i = \text{arg max}_{j} \mathbf{V}_{i,j} \)} & 
        \cellcontent{\( y = \sum_{j=0}^{\text{in\_dim}-1} 1_{\{\mathrm{Sigmoid}(\mathbf{S}_{j} / \tau) \,\geq\, \theta_{\mathrm{sum\text{-}th}}\}} \cdot \mathbf{C}_{j} \cdot \mathbf{x}_j \)} \\
        \bottomrule
    \end{tabular}
    }
\end{table}

\begin{figure}[!htbp]
\centering
\includegraphics[width=0.7\textwidth]{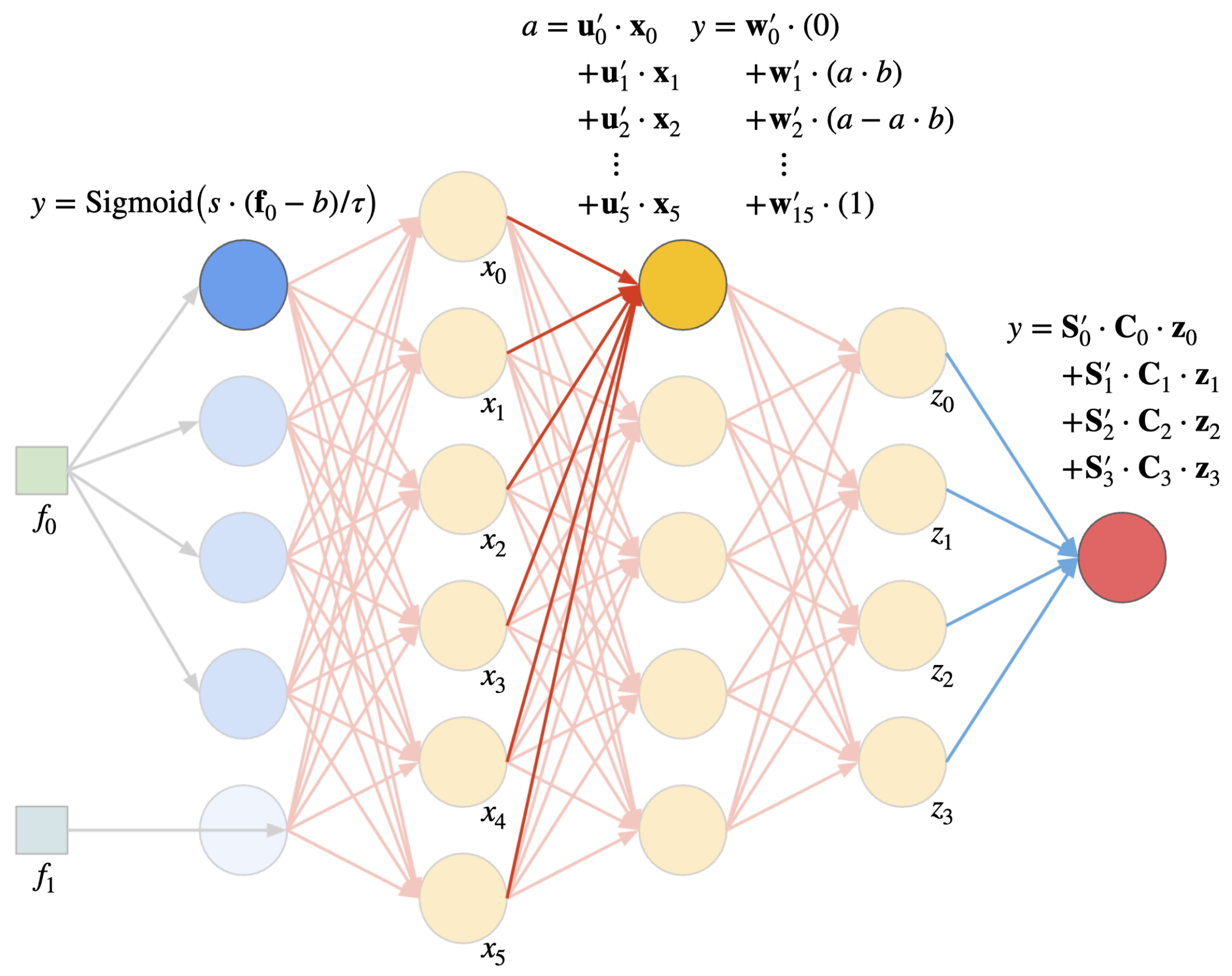}
\caption{Illustration of the training process. The network learns neuron functions and connections
simultaneously: Key details are highlighted for clarity.}
\label{training}
\Description{A diagram illustrating how a network learns during training, 
showing how both neuron functions and their connections are updated at the same time.}
\end{figure}

Table~\ref{layer-params} summarizes the trainable parameters for each layer type and explains their roles during
training and inference.  
% For clarity, we provide the forward-computation equations at the neuron level—namely,
% how the \(i^{\text{th}}\) output \( \mathbf{y}_i \) is computed from the input vector \( \mathbf{x}\).
Fig.~\ref{training} illustrates the differentiable training process, in which neuron functions and connections are
optimized simultaneously. Algorithms~\ref{algo-train} and~\ref{algo-inf} detail the training and inference 
procedures, respectively.

% \textit{ThresholdLayer.}  During inference, this layer employs the Heaviside step function; during training, it uses a
% scaled and shifted sigmoid to preserve differentiability.

% \textit{LogicLayer.}  This layer learns both the Boolean operation performed by each neuron and the two incoming
% connections.  Each candidate function or connection is associated with a learnable logit interpreted, after a
% softmax, as a probability.  After training, every logit is quantized to its highest-probability discrete
% choice, yielding a fully logical circuit.

% \textit{SumLayer.}  This layer contains (i) binary connection weights that specify which neurons in the previous layer
% contribute to the output and (ii) continuous coefficients that weight those contributions.  At inference time, the
% connection weights are binarized, whereas the coefficients remain in floating-point format.

% Fig.~\ref{training} illustrates the differentiable training process, in which neuron functions and connections are
% optimized simultaneously.

\begin{algorithm}[tb]
\centering
\caption{Training}\label{algo-train}
\begin{algorithmic}[1]
    \REQUIRE Training dataset \( \mathcal{D} \), initial temperature \(\tau\), temperature decay rate \(\gamma\), minimum temperature \(\tau_{\text{min}}\), SumLayer link threshold \( {\theta}_{\text{sum-th}} \)
    \ENSURE Trained model parameters \( \mathbf{b}, \mathbf{s}, \mathbf{W}, \mathbf{U}, \mathbf{V}, \mathbf{S}, \mathbf{C} \)
    \FOR{\( \text{epoch} = 1, 2, \ldots, E \)}
        \FORALL{\( (\mathbf{x}, z) \) in \( \mathcal{D} \)}
            \STATE \COMMENT{Forward pass}
            \FOR{layer in model\_layers}
                \IF{\( \text{layer} = \text{ThresholdLayer} \)} \label{algo-train-thresh-start}
                    \STATE \COMMENT {Trainable parameters: \( \mathbf{b}, \mathbf{s} \)}
                    \STATE \( \mathbf{y}_i = \text{Sigmoid}(\mathbf{s}_i \cdot (\mathbf{x}_i - \mathbf{b}_i)) \) \label{algo-train-thresh-end}
                \ELSIF{\( \text{layer} = \text{LogicLayer} \)}
                    \STATE \COMMENT {Trainable parameters: \( \mathbf{W}, \mathbf{U}, \mathbf{V}\)}
                    \STATE \( \mathbf{P} = \text{Softmax}(\mathbf{W}_{i,:} / \tau) \) \label{algo-train-logic-prob}
                    \STATE \( a = \sum_{j=0}^{\text{in\_dim}-1} \left[ \text{Softmax}(\mathbf{U}_{i,:} / \tau) \right]_j \cdot \mathbf{x}_j \) \label{algo-train-a}
                    \STATE \( b = \sum_{j=0}^{\text{in\_dim}-1} \left[ \text{Softmax}(\mathbf{V}_{i,:} / \tau) \right]_j \cdot \mathbf{x}_j \) \label{algo-train-b}
                    \STATE \( \mathbf{y}_i = \sum_{k=0}^{15} \mathbf{P}_k \cdot \text{SoftLogic}_k (a, b) \) \label{algo-train-softlogic} \label{algo-train-logic-sum}
                \ELSE
                    \STATE \COMMENT {Trainable parameters: \( \mathbf{S}, \mathbf{C} \)}
                    \STATE \( \mathbf{y} = \sum_{j=0}^{\text{in\_dim}-1} \text{Sigmoid}(\mathbf{S}_{j} / \tau) \cdot \mathbf{C}_{j} \cdot \mathbf{x}_j \) \label{algo-train-sum}
                \ENDIF
                \STATE \( \mathbf{x} = \mathbf{y} \) \COMMENT {Take preceding layer's output as input}
            \ENDFOR
            \STATE{} \COMMENT{Backward pass}
            \STATE Compute loss \( \mathcal{L}(\mathbf{x}, z) \)
            \STATE Backpropagate to compute gradients
            \STATE Update neuron function parameters \( \mathbf{b}, \mathbf{s}, \mathbf{W} \)
            \STATE \( \tau = \text{max} ( \tau \times \gamma, \tau_{\text{min}}) \) \COMMENT {Temperature decay}
        \ENDFOR
    \ENDFOR
\end{algorithmic}
\end{algorithm}

\begin{algorithm}[tb]
\centering
\caption{Inference}\label{algo-inf}
\begin{algorithmic}[1]
    \REQUIRE Input data \( \mathbf{x} \)
    \REQUIRE Model layer list \( \left[ \begin{aligned} &\text{ThresholdLayer} \\
                                                        & \text{LogicLayer}_1 \\
                                                        & \ldots \\
                                                        & \text{LogicLayer}_n \\
                                                        &\text{SumLayer} \end{aligned} \right] \)
    \STATE \COMMENT{First layer: ThresholdLayer}
    \STATE \( \mathbf{y}_i = \text{Heaviside}(\mathbf{s}_i \cdot (\mathbf{x}_i - \mathbf{b}_i)) \) \label{algo-inf-thresh}
    \STATE \COMMENT {Middle layers: LogicLayer}
    \FOR {\( l = 1, 2, \ldots, n \)}
        \STATE \( k = \text{arg max}_{j} \, \mathbf{W}_{i,j} \) \label{algo-inf-logic-idx}
        \STATE \( a_i = \text{arg max}_{j} \, \mathbf{U}_{i,j} \) \label{algo-inf-a}
        \STATE \( b_i = \text{arg max}_{j} \, \mathbf{V}_{i,j} \) \label{algo-inf-b}
        \STATE \( \mathbf{x} = \mathbf{y} \)  \COMMENT {Take preceding layer's output as input}
        \STATE \( \mathbf{y}_i = \text{HardLogic}_k (\mathbf{x}_{a_i}, \mathbf{x}_{b_i}) \) \label{algo-inf-hardlogic} \label{algo-inf-logic-op}
    \ENDFOR
    \STATE \COMMENT {Last layer: SumLayer}
    \STATE \( \hat{y} = \sum_{j=0}^{\text{in\_dim}-1} 1_{\{ \text{Sigmoid}(\mathbf{S}_{j} / \tau) \ge {\theta}_{\text{sum-th}} \}} \cdot \mathbf{C}_{j} \cdot \mathbf{y}_j \) \label{algo-inf-sum}
    \RETURN \( \hat{y} \)
\end{algorithmic}
\end{algorithm}

% Algorithms~\ref{algo-train} and~\ref{algo-inf} detail the training and inference procedures, respectively.  For
% illustration, the pseudo-code assumes a mini-batch size of one and tracks how each neuron's output \( \mathbf{y}_i \) is
% generated at every layer.  In practice, we train with a batch size of~32.  Because the forward pass of every layer is
% differentiable, gradient-based back-propagation proceeds automatically.  We present additional mathematical details in the 
% following subsections.

\subsection{Unified End-to-End Optimization}
% We propose a single end-to-end optimization procedure for regression DLNs that simultaneously learns
% (i) the threshold parameters in the \textit{ThresholdLayer}; (ii) the logical functions of neurons in the
% \textit{LogicLayer}; (iii) the connection structure in both the \textit{LogicLayer} and the \textit{SumLayer}; and
% (iv) the rule coefficients in the \textit{SumLayer}.  Unlike prior work that splits these steps into separate
% training phases, our unified method updates all trainable parameters at once through continuous relaxations.
% Fig.~\ref{training} highlights the main components of this process.  Next, we describe each layer's formulation and
% objective.
Our unified method updates all trainable parameters (thresholds, logic functions, connection structures, and rule coefficients) simultaneously through continuous relaxations. We describe each layer's formulation below.

\subsubsection{ThresholdLayer}
This layer converts continuous inputs to binary features through a differentiable approximation of the
Heaviside step.  During training, each threshold neuron with input \(x\) has a learnable bias \(b\) and slope \(s\)
and outputs
\[
\begin{aligned}
f(x) &= \text{Sigmoid} (s \cdot (x - b) /\tau ) \\
&= \frac{1}{1 + e^{- [s \cdot (x - b) /\tau ]}},
\end{aligned}
\]
where \(\tau\) is a temperature parameter shared across all sigmoid and softmax operations.  At
inference time, the neuron is discretized as
\[
F(x) = \text{Heaviside} (s \cdot (x - b))
\]
yielding a strict binary output.  Temperature \(\tau\) is annealed during training—an approach that markedly improves
convergence for regression DLNs.  A single threshold neuron's forward pass is depicted in Fig.~\ref{training}. The
corresponding pseudo-code appears in lines~\ref{algo-train-thresh-start}–\ref{algo-train-thresh-end} of
Algorithm~\ref{algo-train}.  The inference step is shown in line~\ref{algo-inf-thresh} of
Algorithm~\ref{algo-inf}.

\textit{Data preprocessing.}  Following the original DLN study, we apply min-max scaling to continuous variables and
one-hot encoding to categorical variables so that every input lies in \([0,1]\).  Whereas classification DLNs allocate
four or six threshold neurons per continuous feature, we find regression tasks benefit from greater resolution and
use six or ten neurons instead.

\textit{Initialization.}  We set all slopes to \(s = 2\), giving the sigmoids enough smoothness for informative
gradients early in training, and initialize biases \(b\) from the bin edges of a decision-tree regressor fitted to
each feature.  After training, any bias that converges outside \([0,1]\) is interpreted as a constant true or
false, depending on the sign of \(s\).

\subsubsection{LogicLayer}
Following LGN~\cite{petersen2022deep} and DLN~\cite{10681646}, we employ \emph{real-valued logic} to sidestep the
non-differentiability of Boolean operators.  As summarized in Table I of DLN~\cite{10681646}, real-valued logic
extends Boolean logic by mapping both inputs and outputs to the interval \([0,1]\).  We write
\(\mathrm{SoftLogic}_k\) for the differentiable form of the \(k^{\text{th}}\) operator and
\(\mathrm{HardLogic}_k\) for its Boolean counterpart.  For example, the real-valued \textsc{and} gate is
\[
\begin{aligned}
\text{real-valued AND} (a, b) &= \text{SoftLogic}_1 (a, b) \\
&= a \cdot b,
\end{aligned}
\]
where \( a, b \in [0, 1] \) are continuous real numbers. The Boolean gate used at inference time is
\[
\begin{aligned}
\text{binary AND} (a, b) &= \text{HardLogic}_1 (a, b) \\
&= 1_{\{ a = 1\}} \cdot 1_{\{ b = 1\}},
\end{aligned}
\]
where \( a, b \in \{0, 1\} \) are discrete binary numbers and indicator functions are applied to them. 
During training, we use \(\mathrm{SoftLogic}_k\) (Algorithm \ref{algo-train}, line~\ref{algo-train-softlogic}); at
inference time, we switch to \(\mathrm{HardLogic}_k\) (Algorithm \ref{algo-inf}, line~\ref{algo-inf-hardlogic}).

Each LogicLayer neuron realizes one of the 16 two-input Boolean functions.  Following LGN~\cite{petersen2022deep}, we represent the
function choice with a learnable logit vector \(\mathbf w\in\mathbb R^{16}\).  Given scalar inputs \(a\) and \(b\),
the neuron's output during training is
\[
\begin{aligned}
y &= \sum_{k=0}^{15} \Bigl[ \text{Softmax}(\mathbf{w} /\tau) \Bigr]_k \cdot \text{SoftLogic}_k (a, b) \\
&= \sum_{k=0}^{15} \frac{e^{\mathbf{w}_k /\tau}}{\sum_j e^{\mathbf{w}_j /\tau}} \cdot \text{SoftLogic}_k (a, b),
\end{aligned}
\]
where \(\tau\) is the global temperature.

Selecting the two incoming signals is another discrete choice.  Each neuron therefore maintains two logit vectors
\(\mathbf u,\mathbf v\in\mathbb R^{\text{in\_dim}}\), whose softmaxes pick weighted combinations of the previous
layer's outputs \(\mathbf x\):
\[
\begin{aligned}
a &= \sum_{j=0}^{\text{in\_dim}-1} \Bigl[\mathrm{Softmax}(\mathbf{u} / \tau)\Bigr]_j \cdot \mathbf{x}_j, \\
b &= \sum_{j=0}^{\text{in\_dim}-1} \Bigl[\mathrm{Softmax}(\mathbf{v} / \tau)\Bigr]_j \cdot \mathbf{x}_j.
\end{aligned}
\]
These steps appear in Algorithm \ref{algo-train}, lines~\ref{algo-train-logic-prob}–\ref{algo-train-logic-sum}, and
are highlighted for one neuron in Fig.~\ref{training}.

At inference time, we quantize to the arg-max choices:
\[
\begin{aligned}
k &= \arg\max_{i} \, \mathbf{w}_i, \\
a_i &= \arg\max_{i} \, \mathbf{u}_{i}, \\
b_i &= \arg\max_{i} \, \mathbf{v}_{i}, \\
\end{aligned}
\]
\[
y = \mathrm{HardLogic}_k\bigl(\mathbf{x}_{a_i}, \mathbf{x}_{b_i}\bigr).
\]
The corresponding pseudo-code is given in Algorithm \ref{algo-inf},
lines~\ref{algo-inf-logic-idx}–\ref{algo-inf-hardlogic}.

\subsubsection{SumLayer}
The SumLayer in the regression DLN diverges markedly from its counterpart in classification DLNs.  Because a
regression model outputs a single continuous value, the SumLayer computes a weighted sum of the activations from the
preceding (final) LogicLayer. 
We test several variants for this layer, such as L1-regularized, 
but find this vanilla weighted-sum version performs the best.
We leave the coefficients \(\mathbf c\) for these activations in floating-point
format; only the binary connectivity pattern is later quantized.  As in the original DLN, we represent the presence
of each potential input connection with a learnable sigmoid gate.  Concretely, for every candidate input
\(\mathbf x_j\), the layer maintains (i) a link-strength logit \(\mathbf s_j\in\mathbb R\) and (ii) a
continuous rule weight \(\mathbf c_j\in\mathbb R\).  Given the previous layer's output vector \(\mathbf x\), during training, the 
layer outputs 
\[
y = \sum_{j=0}^{\text{in\_dim}-1} \mathrm{Sigmoid}\bigl(\mathbf{s}_{j} / \tau \bigr) \cdot \mathbf{c}_{j} \cdot \mathbf{x}_j.
\]
After training, we discretize the connection pattern by retaining links whose gate value exceeds a fixed threshold
\(\theta_{\text{sum-th}}\).  The inference computation therefore becomes
\[
y = \sum_{j=0}^{\text{in\_dim}-1} 1_{\{\mathrm{Sigmoid}(\mathbf{s}_{j} / \tau) \ge \theta_{\mathrm{sum\text{-}th}}\}} \cdot \mathbf{c}_{j} \cdot \mathbf{x}_j.
\]
where we set \(\theta_{\text{sum-th}}=0.8\), following DLN.  The corresponding pseudo-code appears in
line~\ref{algo-train-sum} of Algorithm~\ref{algo-train} and line~\ref{algo-inf-sum} of
Algorithm~\ref{algo-inf}.

\subsection{Training Strategies and Model Simplification}\label{method-strategy}
To train regression DLNs effectively, we adapt several strategies originally introduced for classification
DLNs~\cite{10681646}.  We summarize these methods below and apply them analogously in our framework; readers seeking
additional detail can consult the cited work.

\begin{itemize}
\item \textbf{Searching over Subspaces:}  To curb the size of the search space for neuron functions and input
      links, we restrict each neuron to a subset of candidate gates and connections.  We set the logits of excluded
      options to \(-\infty\) inside the softmax, thereby focusing optimization on a manageable subspace and
      improving gradient flow.

\item \textbf{Using Straight-Through Estimators (STEs):}  Cascaded sigmoids and softmax operations
      can dull activations.  We employ the STE~\cite{bengio2013estimating} so that the forward pass uses discrete
      outputs while the backward pass propagates gradients from their continuous relaxations.  This adjustment
      makes regression DLNs substantially more trainable.

\item \textbf{Concatenating Inputs:}  Inspired by the Wide \& Deep model~\cite{cheng2016wide} and DLN, we
      concatenate the binarized inputs produced by the ThresholdLayer to the inputs of intermediate LogicLayers.
      This shortcut improves information flow, grants later layers direct access to raw features, and eases
      gradient propagation.

\item \textbf{Model Simplification:}  After training, we extract the learned logical expressions and simplify them
      with symbolic tools (e.g., SymPy~\cite{10.7717/peerj-cs.103}), mirroring the procedure in DLN.  This step reduces model size and
      enhances interpretability.
\end{itemize}

We validate the effectiveness of these strategies through ablation studies whose results appear in
Sec.~\ref{exp-ablation}.

\textbf{Temperature Scheduling.}  Unlike the original DLN, our training employs an exponential temperature decay
for every sigmoid and softmax.  The temperature \(\tau\) is initialized to a large value and
multiplied by a factor \(\gamma<1\) after each epoch until it reaches a minimum \(\tau_{\text{min}}\).  The trio
\((\tau,\gamma,\tau_{\text{min}})\) is treated as a set of hyperparameters and tuned during
experimentation.

\section{Experiments}\label{sec-experiments}
\begin{figure}[!htbp]
\centering
\includegraphics[width=0.75\columnwidth]{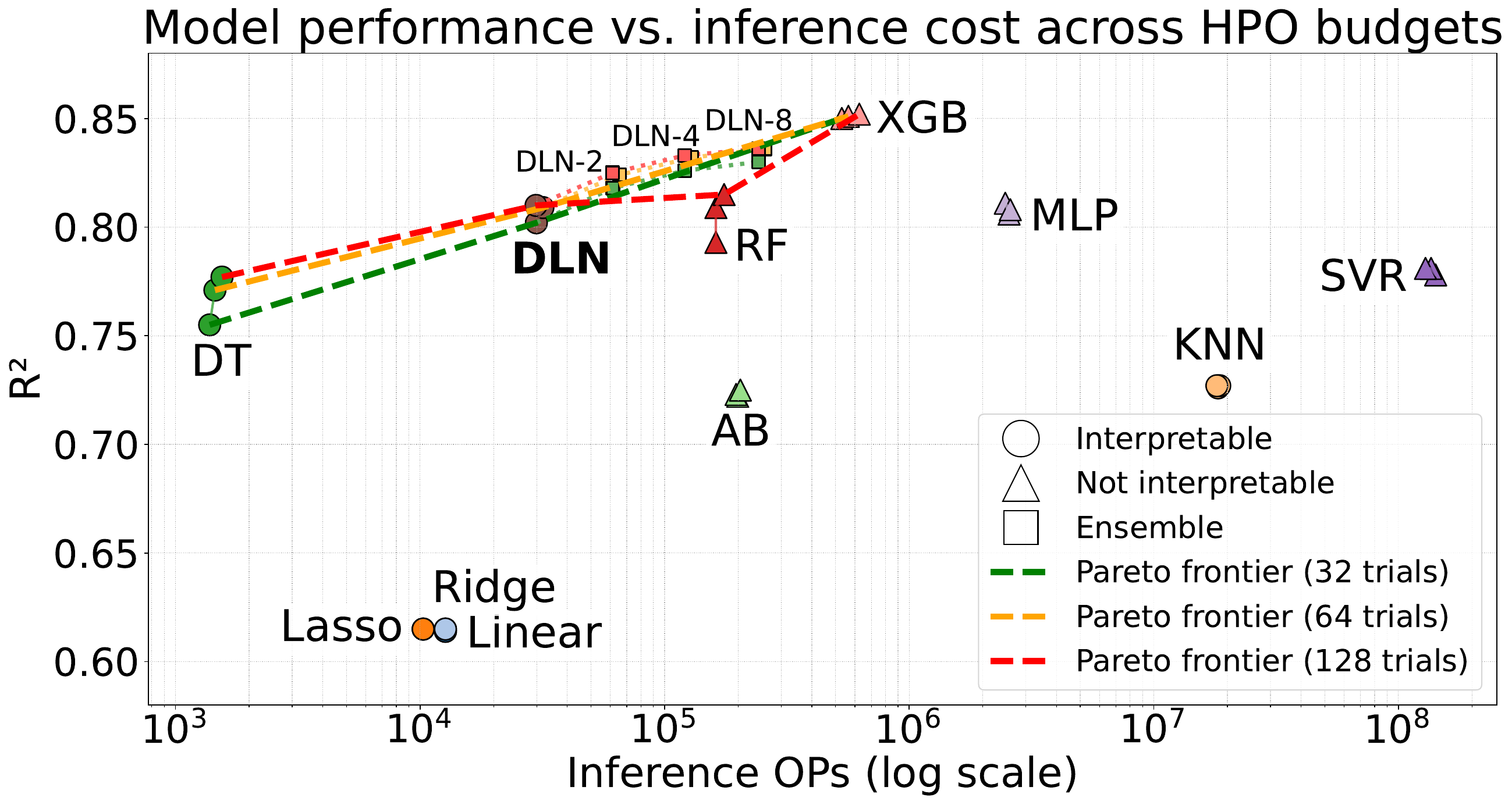}
\caption{Comparison of $R^{2}$ and the number of operations required for inference across models.  We plot results
for 32, 64, and 128 hyperparameter-search trials and draw the Pareto frontier for each case.  DLN is Pareto-optimal in
all three. XGBoost is the best-performing model in terms of accuracy. The vanilla DLN ensembles extend the Pareto 
frontier, outperforming random forest, a model that uses a similar ensembling framework, on both accuracy and inference cost.}
\label{pareto}
\Description{A scatter plot comparing models based on their R-squared score versus 
the number of inference operations. It displays Pareto frontiers for different 
trial counts, showing that the DLN model is consistently on the optimal frontier.}
\end{figure}

We evaluate the proposed DLN and its three DLN ensemble variants against 10 baseline methods on 15 tabular regression datasets drawn from the
UC Irvine Machine Learning Repository~\cite{ucirepo} and the Kaggle platform~\cite{kaggle}.  The baselines are linear
regression, Ridge, Lasso, $k$-nearest neighbors (KNN), decision tree (DT), AdaBoost (AB), random forest (RF), 
XGBoost (XGB), support vector regression (SVR), and MLP. 
For each (dataset, model) pair, we train 10 independent models with different random seeds. 
The DLN ensembles, DLN-2, DLN-4, and DLN-8, are formed by averaging the predictions of 2, 4, and 8 models, 
respectively, randomly sampled from these 10 trained DLNs. 
Table~\ref{datasets-stats} lists each dataset's
characteristics after preprocessing; sample sizes range from hundreds to thousands.  We assess predictive accuracy
and inference cost, and we illustrate DLN's interpretability by visualizing its decision process.  As summarized in
Fig.~\ref{pareto}, DLN lies on the Pareto frontier under every experimental setting, demonstrating a strong balance of accuracy and efficiency.

\begin{table}[!t]
\centering
\caption{Characteristics of the datasets after preprocessing}
\label{datasets-stats}
\small{
\begin{tabular}{lrrrr|rl}
\toprule
Dataset & \# Train & \# Test & \# Cont. & \# Cate. & Source & ID \\
\midrule
Abalone     &  3132 &  1045 &  7  &  2  & UCI    & 1 \\ \addlinespace[1pt] 
Airfoil     &  1127 &   376 &  5  &  0  & UCI    & 291 \\ \addlinespace[1pt] 
Bike        & 13032 &  4345 &  6  & 14  & UCI    & 275 \\ \addlinespace[1pt] 
CCPP        &  7145 &  2382 &  4  &  0  & UCI    & 294 \\ \addlinespace[1pt] 
Concrete    &   753 &   252 &  8  &  0  & UCI    & 165 \\ \addlinespace[1pt] 
Electrical  &  7500 &  2500 & 12  &  0  & UCI    & 471 \\ \addlinespace[1pt] 
Energy      &   576 &   192 &  6  &  4  & UCI    & 242 \\ \addlinespace[1pt] 
Estate      &   310 &   104 &  6  &  0  & UCI    & 477 \\ \addlinespace[1pt] 
Housing     & 15324 &  5109 &  8  &  4  & Kaggle & camnugent/california-housing-prices \\ \addlinespace[1pt] 
Infrared    &   763 &   255 & 31  &  6  & UCI    & 925 \\ \addlinespace[1pt] 
Insurance   &  1002 &   335 &  2  & 12  & Kaggle & mirichoi0218/insurance \\ \addlinespace[1pt] 
MPG         &   294 &    98 &  6  &  2  & UCI    & 9 \\ \addlinespace[1pt] 
Parkinson's  &  4406 &  1469 & 18  &  1  & UCI    & 189 \\ \addlinespace[1pt] 
Wine        &  3988 &  1330 & 11  &  0  & UCI    & 186 \\ \addlinespace[1pt] 
Yacht       &   231 &    77 &  6  &  0  & Kaggle & heitornunes/yacht-hydrodynamics-data-set \\ 
\bottomrule
\end{tabular}
}
\end{table}

\subsection{Experimental Setup}
We preprocess each dataset as follows:
\begin{enumerate}
\item Remove rows with missing values.  The selected datasets are almost complete; hence, only a negligible number of
      rows are discarded.
\item One-hot encode categorical features.
\item Min–Max scale continuous features to the \([0,1]\) range.
\item Standardize targets to zero mean and unit variance.
\end{enumerate}

Each of the 10 trials per model 
comprises three stages: hyperparameter search, training, and evaluation.  During hyperparameter search, we employ
Optuna~\cite{optuna_2019} to sample hyperparameter values from a predefined grid and select the value configuration that yields
the lowest cross-validated mean-squared error (MSE).  Depending on the dataset size, we use two to four folds, with
smaller datasets receiving more folds.  Data processing and all classical baselines rely on
scikit-learn~\cite{scikit-learn}. XGBoost is implemented using its official package~\cite{chen2016xgboost}. 
The MLP and DLN models are implemented in PyTorch~\cite{NEURIPS2019_9015}. 
Ray~\cite{moritz2018ray} orchestrates parallel hyperparameter searches and training runs.

\begin{table}[!htbp]
\centering
\caption{Average test $R^{2}$ across 10 random seeds and 128 hyperparameter trials per model}
\label{results-r2}
{\setlength{\tabcolsep}{1.5pt}\small
\begin{tabularx}{\textwidth}{l*{11}{Y}}
\toprule
& Linear & Ridge & Lasso & KNN & DT & AB & RF & XGB & SVR & MLP & DLN \\
\midrule
Abalone & \meanstd{0.534}{0.024} & \meanstd{0.534}{0.024} & \meanstd{0.534}{0.024} & \meanstd{0.517}{0.013} & \meanstd{0.475}{0.018} & \meanstd{0.463}{0.027} & \meanstd{0.547}{0.019} & \meanstd{0.554}{0.018} & \meanstd{0.551}{0.018} & \meanstd{0.570}{0.024} & \meanstd{0.526}{0.026} \\
\addlinespace[2pt]
Airfoil & \meanstd{0.516}{0.034} & \meanstd{0.516}{0.033} & \meanstd{0.516}{0.033} & \meanstd{0.874}{0.026} & \meanstd{0.793}{0.052} & \meanstd{0.743}{0.020} & \meanstd{0.868}{0.016} & \meanstd{0.952}{7.4e-03} & \meanstd{0.822}{0.024} & \meanstd{0.943}{5.7e-03} & \meanstd{0.889}{0.013} \\
\addlinespace[2pt]
Bike & \meanstd{0.403}{0.012} & \meanstd{0.403}{0.012} & \meanstd{0.403}{0.012} & \meanstd{0.659}{0.012} & \meanstd{0.904}{0.011} & \meanstd{0.660}{0.017} & \meanstd{0.870}{7.1e-03} & \meanstd{0.956}{3.2e-03} & \meanstd{0.645}{0.014} & \meanstd{0.939}{3.0e-03} & \meanstd{0.916}{6.4e-03} \\
\addlinespace[2pt]
CCPP & \meanstd{0.929}{2.4e-03} & \meanstd{0.929}{2.4e-03} & \meanstd{0.929}{2.4e-03} & \meanstd{0.957}{2.0e-03} & \meanstd{0.942}{2.8e-03} & \meanstd{0.920}{3.8e-03} & \meanstd{0.947}{4.6e-03} & \meanstd{0.966}{1.3e-03} & \meanstd{0.946}{2.2e-03} & \meanstd{0.948}{2.4e-03} & \meanstd{0.943}{2.3e-03} \\
\addlinespace[2pt]
Concrete & \meanstd{0.593}{0.034} & \meanstd{0.593}{0.034} & \meanstd{0.593}{0.034} & \meanstd{0.709}{0.038} & \meanstd{0.795}{0.029} & \meanstd{0.791}{0.015} & \meanstd{0.877}{0.018} & \meanstd{0.931}{0.010} & \meanstd{0.869}{0.014} & \meanstd{0.882}{0.022} & \meanstd{0.888}{0.020} \\
\addlinespace[2pt]
Electrical & \meanstd{0.647}{0.014} & \meanstd{0.647}{0.014} & \meanstd{0.647}{0.013} & \meanstd{0.798}{4.0e-03} & \meanstd{0.755}{0.012} & \meanstd{0.826}{8.4e-03} & \meanstd{0.843}{0.013} & \meanstd{0.955}{2.7e-03} & \meanstd{0.962}{1.7e-03} & \meanstd{0.968}{2.0e-03} & \meanstd{0.936}{3.6e-03} \\
\addlinespace[2pt]
Energy & \meanstd{0.911}{7.5e-03} & \meanstd{0.911}{7.5e-03} & \meanstd{0.911}{7.4e-03} & \meanstd{0.956}{9.0e-03} & \meanstd{0.995}{2.4e-03} & \meanstd{0.969}{4.7e-03} & \meanstd{0.996}{2.0e-03} & \meanstd{0.998}{2.9e-04} & \meanstd{0.991}{1.2e-03} & \meanstd{0.997}{3.0e-04} & \meanstd{0.998}{1.8e-04} \\
\addlinespace[2pt]
Estate & \meanstd{0.614}{0.052} & \meanstd{0.615}{0.051} & \meanstd{0.614}{0.050} & \meanstd{0.707}{0.062} & \meanstd{0.686}{0.092} & \meanstd{0.716}{0.037} & \meanstd{0.762}{0.036} & \meanstd{0.763}{0.029} & \meanstd{0.720}{0.058} & \meanstd{0.634}{0.065} & \meanstd{0.679}{0.061} \\
\addlinespace[2pt]
Housing & \meanstd{0.661}{9.3e-03} & \meanstd{0.661}{9.3e-03} & \meanstd{0.660}{9.2e-03} & \meanstd{0.745}{6.7e-03} & \meanstd{0.740}{8.3e-03} & \meanstd{0.589}{8.4e-03} & \meanstd{0.747}{0.016} & \meanstd{0.849}{6.8e-03} & \meanstd{0.778}{7.8e-03} & \meanstd{0.796}{9.0e-03} & \meanstd{0.766}{0.014} \\
\addlinespace[2pt]
Infrared & \meanstd{0.736}{0.035} & \meanstd{0.740}{0.035} & \meanstd{0.741}{0.033} & \meanstd{0.692}{0.043} & \meanstd{0.734}{0.035} & \meanstd{0.761}{0.033} & \meanstd{0.769}{0.031} & \meanstd{0.771}{0.028} & \meanstd{0.745}{0.029} & \meanstd{0.667}{0.069} & \meanstd{0.725}{0.040} \\
\addlinespace[2pt]
Insurance & \meanstd{0.738}{0.015} & \meanstd{0.739}{0.015} & \meanstd{0.739}{0.015} & \meanstd{0.698}{0.031} & \meanstd{0.849}{0.019} & \meanstd{0.849}{0.015} & \meanstd{0.858}{0.018} & \meanstd{0.862}{0.015} & \meanstd{0.822}{0.018} & \meanstd{0.806}{0.025} & \meanstd{0.855}{0.013} \\
\addlinespace[2pt]
MPG & \meanstd{0.803}{0.022} & \meanstd{0.803}{0.023} & \meanstd{0.802}{0.024} & \meanstd{0.867}{0.025} & \meanstd{0.818}{0.039} & \meanstd{0.836}{0.025} & \meanstd{0.866}{0.020} & \meanstd{0.870}{0.015} & \meanstd{0.884}{0.021} & \meanstd{0.846}{0.060} & \meanstd{0.856}{0.030} \\
\addlinespace[2pt]
Parkinsons & \meanstd{0.185}{0.020} & \meanstd{0.185}{0.020} & \meanstd{0.185}{0.020} & \meanstd{0.662}{0.012} & \meanstd{0.920}{0.028} & \meanstd{0.462}{0.019} & \meanstd{0.913}{6.9e-03} & \meanstd{0.975}{4.3e-03} & \meanstd{0.614}{0.021} & \meanstd{0.836}{0.058} & \meanstd{0.859}{0.033} \\
\addlinespace[2pt]
Wine & \meanstd{0.312}{0.017} & \meanstd{0.311}{0.017} & \meanstd{0.312}{0.017} & \meanstd{0.358}{0.014} & \meanstd{0.261}{0.019} & \meanstd{0.303}{0.016} & \meanstd{0.362}{0.013} & \meanstd{0.387}{0.011} & \meanstd{0.378}{0.014} & \meanstd{0.290}{0.029} & \meanstd{0.321}{0.020} \\
\addlinespace[2pt]
Yacht & \meanstd{0.631}{0.036} & \meanstd{0.633}{0.034} & \meanstd{0.645}{0.024} & \meanstd{0.706}{0.049} & \meanstd{0.993}{3.2e-03} & \meanstd{0.990}{3.7e-03} & \meanstd{0.995}{2.4e-03} & \meanstd{0.998}{1.2e-03} & \meanstd{0.990}{7.4e-03} & \meanstd{0.993}{4.0e-03} & \meanstd{0.997}{1.5e-03} \\
\midrule
Mean $R^{2}$ $\uparrow$ & 0.614 & 0.615 & 0.615 & 0.727 & 0.777 & 0.725 & 0.815 & 0.852 & 0.781 & 0.808 & 0.810 \\ \addlinespace[1pt]
Average Rank & 9.13 & 9.20 & 8.87 & 6.53 & 6.60 & 7.27 & 3.67 & 1.27 & 4.47 & 4.53 & 4.47 \\
\bottomrule
\end{tabularx}
}
\end{table}

\begin{figure}[!htbp]
\centering
\includegraphics[width=0.55\columnwidth]{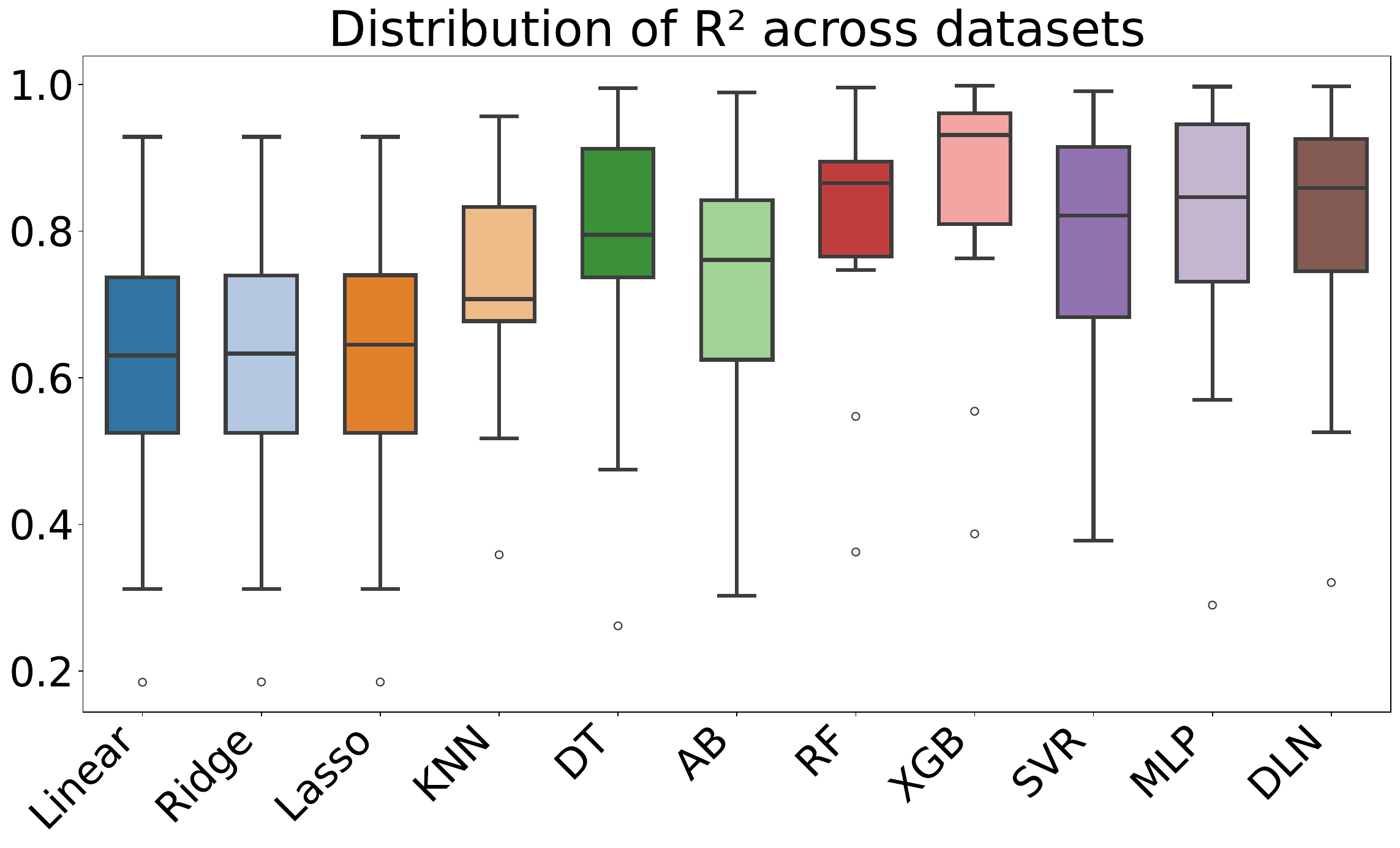}
\caption{Distribution of $R^{2}$ across datasets for each model, averaged over 10 random seeds and 128
hyperparameter-search trials.}
\label{boxplot}
\Description{A box plot showing the distribution of R-squared scores for 
several machine learning models across multiple datasets. The plot 
illustrates the median, quartiles, and range of performance for each model.}
\end{figure}

\subsection{Accuracy}
The performance of the regression models is measured by three main metrics: $R^{2}$, root-mean-square error (RMSE),
and mean absolute error (MAE).  We present the $R^{2}$ results in this section and provide the RMSE and MAE values in
the Appendix.  As expected, $R^{2}$ is strongly and inversely correlated with RMSE and MAE.  Table~\ref{results-r2}
reports the average $R^{2}$ for all models on the 15 datasets. Fig.~\ref{boxplot} shows the full
$R^{2}$ distribution for each model.  DLN achieves the third-highest mean $R^{2}$ and the third-best average rank
across all datasets, with XGBoost ranking first.

XGBoost, a highly-optimized gradient boosting method, achieves the best-performing mean $R^2$ and average rank, 
confirming its state-of-the-art status on tabular data. Our single DLN model achieves accuracy competitive with 
random forest and MLP. Furthermore, we find that DLN benefits significantly from simple ensembling. 
A vanilla mean ensemble of two DLNs (DLN-2) outperforms random forest in both mean $R^2$ and mean inference OP cost. 
It is noteworthy that our current ensemble approach employs simple model averaging, a method theoretically 
inferior to the ``bagging'' mechanism of random forest and the ``boosting'' framework of XGBoost. 
The fact that a simple average of DLNs outperforms the more sophisticated bagging mechanism of random forest 
suggests that the DLN is a highly effective base estimator.
While simple averaging yields diminishing returns and likely cannot surpass sophisticated boosting methods 
like XGBoost in pure accuracy, these results highlight a promising future direction: integrating DLNs as learners 
within a gradient boosting framework. Furthermore, most ensembling techniques involve ensembling weak models to reduce the 
overall inference computation. We currently fuse individual DLNs directly. Thus, ensembled DLNs have the potential 
to reduce inference cost further if we encourage smaller models during HPO. We believe efficient DLN ensembling 
is a promising future direction to explore.

Linear regression and its two regularized variants, Ridge and Lasso, perform similarly but lack sufficient capacity for
many tasks.  KNN substantially improves over the linear models.  Decision trees push the upper bound higher but are
less stable.  The boosting variant, AdaBoost, does not improve on a single tree, whereas the bagging variant, random
forest, is robust—it attains a high peak and performs well on difficult datasets such as Parkinson's and Wine. 
However, it is surpassed by XGBoost, which leverages a more sophisticated boosting framework to set
the new state-of-the-art for accuracy.
Support vector regression also excels on some complex datasets but is less consistent overall. The MLP performs well
on datasets that challenge most other methods (e.g., Airfoil and Bike), demonstrating its strength on highly
nonlinear problems. It attains strong overall accuracy. Our single DLN model's accuracy is competitive with RF and MLP, 
and the DLN-2 ensemble surpasses random forest.

\begin{figure}[!htbp]
\centering
\includegraphics[width=0.55\columnwidth]{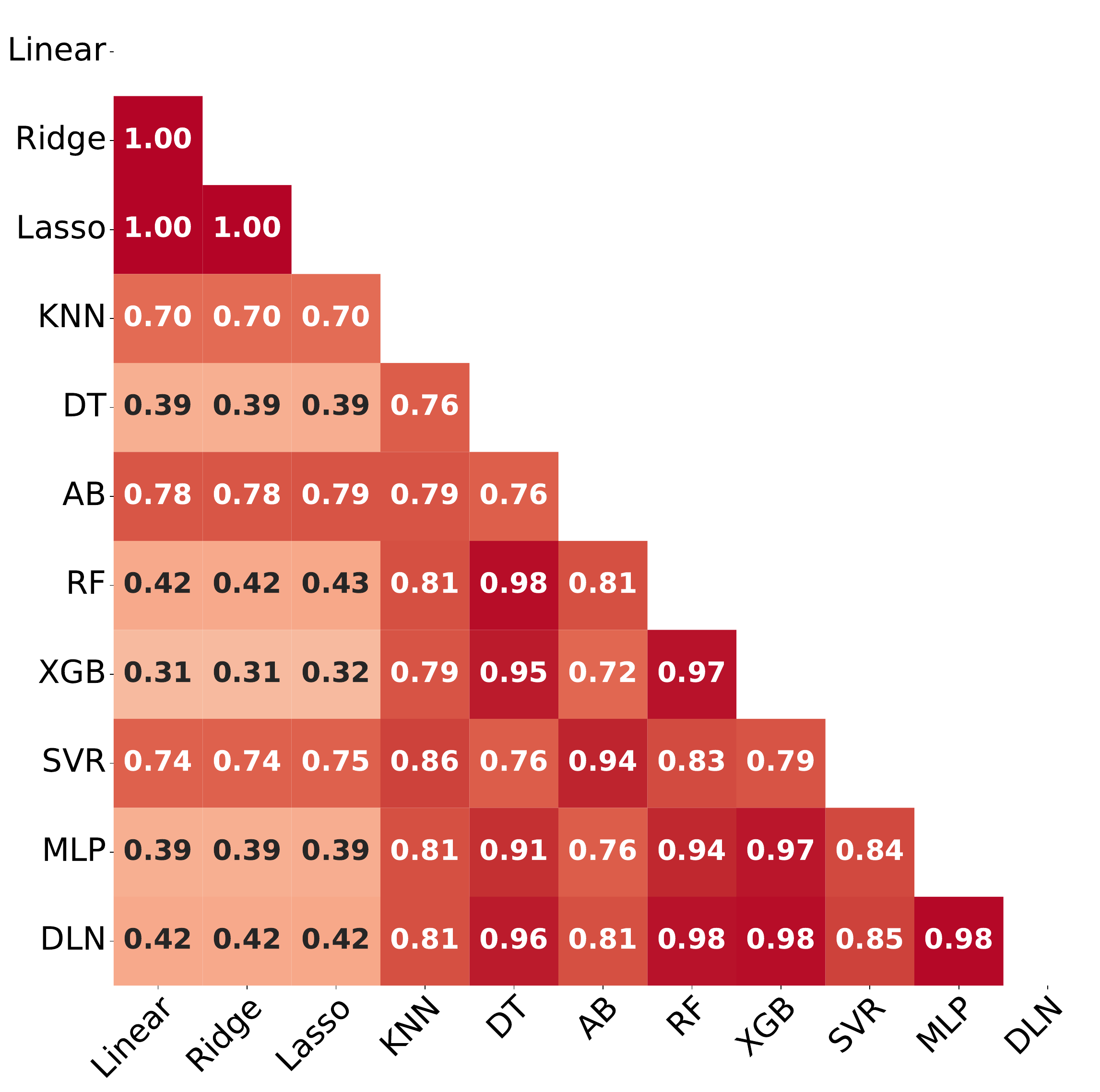}
\caption{Pearson correlation matrix of model $R^{2}$ scores, averaged over 10 random seeds and 128
hyperparameter-search trials.}
\label{correlation}
\Description{A heatmap representing the Pearson correlation matrix of R-squared 
scores among different models. The color of each cell indicates the strength 
and direction of the correlation between two models' performances.}
\end{figure}

We plot the Pearson correlation matrix of model $R^{2}$ scores in Fig.~\ref{correlation}.  A strong correlation is
evident within the linear group (Linear, Ridge, Lasso) and within the tree-based group (DT, RF, XGB).  DLN correlates 
most strongly with the tree-based ensembles (RF, XGB) and the MLP.

Finally, we compare performance under different hyperparameter-search budgets.  Table~\ref{results-hpo} lists each
model's average $R^{2}$ for 32, 64, and 128 trials.  Nonlinear models generally improve with larger budgets; random
forest benefits the most from additional trials.

\begin{table}[!htbp]
\centering
\caption{Mean $R^{2}$ scores across HPO budgets}
\label{results-hpo}
\small{
\begin{tabularx}{0.5\columnwidth}{l@{\hspace{2.5em}} *{3}{L}}
\toprule
 & 32 trials & 64 trials & 128 trials \\
\midrule
Linear & 0.614 & 0.614 $\rightarrow$ & 0.614 $\rightarrow$ \\
Ridge  & 0.615 & 0.615 $\rightarrow$ & 0.615 $\rightarrow$ \\
Lasso  & 0.615 & 0.615 $\rightarrow$ & 0.615 $\rightarrow$ \\
KNN    & 0.726 & 0.727 $\uparrow$    & 0.727 $\rightarrow$ \\
DT     & 0.755 & 0.771 $\uparrow$    & 0.777 $\uparrow$    \\
AB     & 0.722 & 0.723 $\uparrow$    & 0.725 $\uparrow$    \\
RF     & 0.793 & 0.809 $\uparrow$    & 0.815 $\uparrow$    \\
XGB    & 0.850 & 0.851 $\uparrow$    & 0.852 $\uparrow$    \\
SVR    & 0.778 & 0.781 $\uparrow$    & 0.781 $\rightarrow$ \\
MLP    & 0.806 & 0.811 $\uparrow$    & 0.808 $\downarrow$  \\
DLN    & 0.802 & 0.809 $\uparrow$    & 0.810 $\uparrow$    \\
\bottomrule
\end{tabularx}
}
\end{table}

\subsection{Efficiency}
This section evaluates the model's inference efficiency using two different metrics: logic operations and CPU latency.

\subsubsection{Inference Cost}
We follow DLN's efficiency protocol and measure inference cost as the number of basic hardware logic-gate
operations (OPs); results are shown in Table~\ref{results-ops}. This metric is useful for estimating 
performance on FPGAs or ASICs. Each model's computation is first decomposed into 
high-level operations, such as floating-point additions and multiplications, which are then mapped to NOT, AND, OR,
NAND, NOR, XOR, or XNOR gates.  A two-input AND, OR, NAND, or NOR counts as one OP, a two-input XOR or XNOR as three
OPs, and a NOT as zero OPs.  During training, we use scikit-learn's and PyTorch's defaults (float32 for MLP and DLN;
float64 for most traditional models), but for the cost analysis, we assume float16 for floating-point and int16 for
integer arithmetic, since nearly all methods retain accuracy with these precisions.  DLN involves very few
floating-point calculations; hence, its relative advantage would grow if larger numeric formats were required.

From Table~\ref{results-ops}, the decision tree can be seen to be the most efficient model, followed by Lasso regression.
Linear and Ridge regressions are only marginally more expensive than Lasso.  DLN ranks next, with a geometric-mean
inference cost $5.8\times$ lower than random forest, $20\times$ lower than XGBoost, and $86\times$ lower than MLP.

\begin{table}[!htbp]
\centering
\caption{Average number of basic gate-level logic operations required for inference across 10 random seeds and
128 hyperparameter trials, assuming FP16 for floating-point and INT16 for integer arithmetic}
\label{results-ops}
\small{
\begin{tabularx}{\textwidth}{l*{11}{Y}}
\toprule
& Linear & Ridge & Lasso & KNN & DT & AB & RF & XGB & SVR & MLP & DLN \\
\midrule
Abalone & 11.2K & 11.2K & 10.7K & 38.5M & 1.19K & 206K & 240K & 432K & 383M & 829K & 41.2K \\
\addlinespace[2pt]
Airfoil & 6.22K & 6.22K & 6.22K & 9.66M & 2.12K & 295K & 174K & 853K & 109M & 1.71M & 32.6K \\
\addlinespace[2pt]
Bike & 24.9K & 24.9K & 21.9K & 271M & 2.85K & 158K & 177K & 1.01M & 1.30G & 12.9M & 50.0K \\
\addlinespace[2pt]
CCPP & 4.98K & 4.98K & 4.98K & 38.9M & 2.01K & 199K & 201K & 960K & 504M & 1.22M & 22.5K \\
\addlinespace[2pt]
Concrete & 9.95K & 9.95K & 8.71K & 9.46M & 2.05K & 293K & 161K & 678K & 67.6M & 2.57M & 60.6K \\
\addlinespace[2pt]
Electrical & 14.9K & 14.9K & 13.3K & 89.5M & 2.36K & 296K & 191K & 840K & 1.03G & 4.65M & 91.1K \\
\addlinespace[2pt]
Energy & 12.4K & 12.4K & 11.6K & 5.55M & 1.37K & 213K & 161K & 509K & 6.73M & 2.03M & 19.8K \\
\addlinespace[2pt]
Estate & 7.46K & 7.46K & 6.47K & 1.31M & 988 & 275K & 156K & 408K & 28.5M & 962K & 12.3K \\
\addlinespace[2pt]
Housing & 14.9K & 14.9K & 14.9K & 144M & 2.58K & 207K & 207K & 1.07M & 1.42G & 2.14M & 49.5K \\
\addlinespace[2pt]
Infrared & 46.0K & 46.0K & 25.6K & 17.7M & 695 & 174K & 167K & 454K & 39.1M & 21.8M & 29.9K \\
\addlinespace[2pt]
Insurance & 17.4K & 17.4K & 12.9K & 16.2M & 1.15K & 94.9K & 136K & 326K & 13.6M & 801K & 5.76K \\
\addlinespace[2pt]
MPG & 9.95K & 9.95K & 8.96K & 1.78M & 1.35K & 206K & 176K & 582K & 33.6M & 2.28M & 19.6K \\
\addlinespace[2pt]
Parkinsons & 23.6K & 23.6K & 22.3K & 67.5M & 2.51K & 176K & 191K & 631K & 578M & 7.20M & 46.7K \\
\addlinespace[2pt]
Wine & 13.7K & 13.7K & 13.7K & 28.9M & 952 & 206K & 194K & 875K & 502M & 3.30M & 48.6K \\
\addlinespace[2pt]
Yacht & 7.46K & 7.46K & 1.87K & 2.20M & 1.08K & 182K & 122K & 403K & 15.9M & 1.50M & 16.7K \\
\midrule
Geo. Mean OPs $\downarrow$ & 12.7K & 12.7K & 10.3K & 18.1M & 1.55K & 204K & 175K & 625K & 129M & 2.59M & 29.8K \\ \addlinespace[1pt]
Average Rank $\downarrow$ & 3.50 & 3.50 & 2.33 & 9.93 & 1.00 & 6.67 & 6.33 & 8.00 & 10.93 & 9.13 & 4.67 \\
\bottomrule
\end{tabularx}
}
\end{table}

\subsubsection{Inference Latency}
To address the limitations of logic OPs as a metric and to provide a more practical measure of efficiency, 
we conducted inference latency tests on a single-threaded CPU at 3.8 GHz. We compare DLN against MLP and XGBoost, 
as they are highly-optimized for inference and represent top-performing black-box models.

To maximize DLN's performance, we develop a custom code generator to translate the trained PyTorch model into 
an AOT-compiled C shared library. This eliminates Python overhead and optimizes the logical computations. 
We compare this AOT-compiled DLN against the standard PyTorch MLP and the official XGBoost library. 
We test at batch sizes of 1 (common for real-time edge applications) and 32. We omit other scikit-learn models, 
as they are not optimized for low-latency inference; for example, random forest was found to be substantially slower.

We show the results in Table~\ref{results-latency}. To compute these values, we test the latency of each model 
on each (dataset, seed) pair, take the mean across the 10 random seeds for each dataset, and finally take the 
geometric mean of these per-dataset averages across all 15 datasets. DLN achieves the best geometric mean speed 
across all datasets, running on average more than 3$\times$ faster than MLP and more than 18$\times$ faster than 
XGBoost at a batch size of 1.
This demonstrates the practical, real-world efficiency of the DLN architecture, even without specific hardware acceleration. 
We note from the table that when the batch size increases to 32, the MLP's per-sample latency improves significantly 
as it benefits from data parallelization. While DLN remains the fastest model, its speedup is more modest. Investigating 
how to better parallelize DLN's logic-based operations for larger batch sizes is a promising direction for future work.

\begin{table}[!htbp]
\centering
\caption{Per-sample inference latency (in microseconds) on a single-threaded CPU. Values are the geometric 
mean (across 15 datasets) of the per-dataset average latencies (mean over 10 seeds). Lower is better.}
\label{results-latency}
\small{
\begin{tabularx}{0.4\columnwidth}{l Y Y Y}
\toprule
Batch Size & XGB & MLP & DLN \\
\midrule
Batch = 1 & 227 \(\mu\text{s}\) & 39.8 \(\mu\text{s}\) & 12.6 \(\mu\text{s}\) \\
Batch = 32 & 13.8 \(\mu\text{s}\) & 1.62 \(\mu\text{s}\) & 1.11 \(\mu\text{s}\) \\
\bottomrule
\end{tabularx}
}
\end{table}

\subsection{Interpretability}
\begin{figure}[!htbp]
\centering
\includegraphics[width=0.55\textwidth]{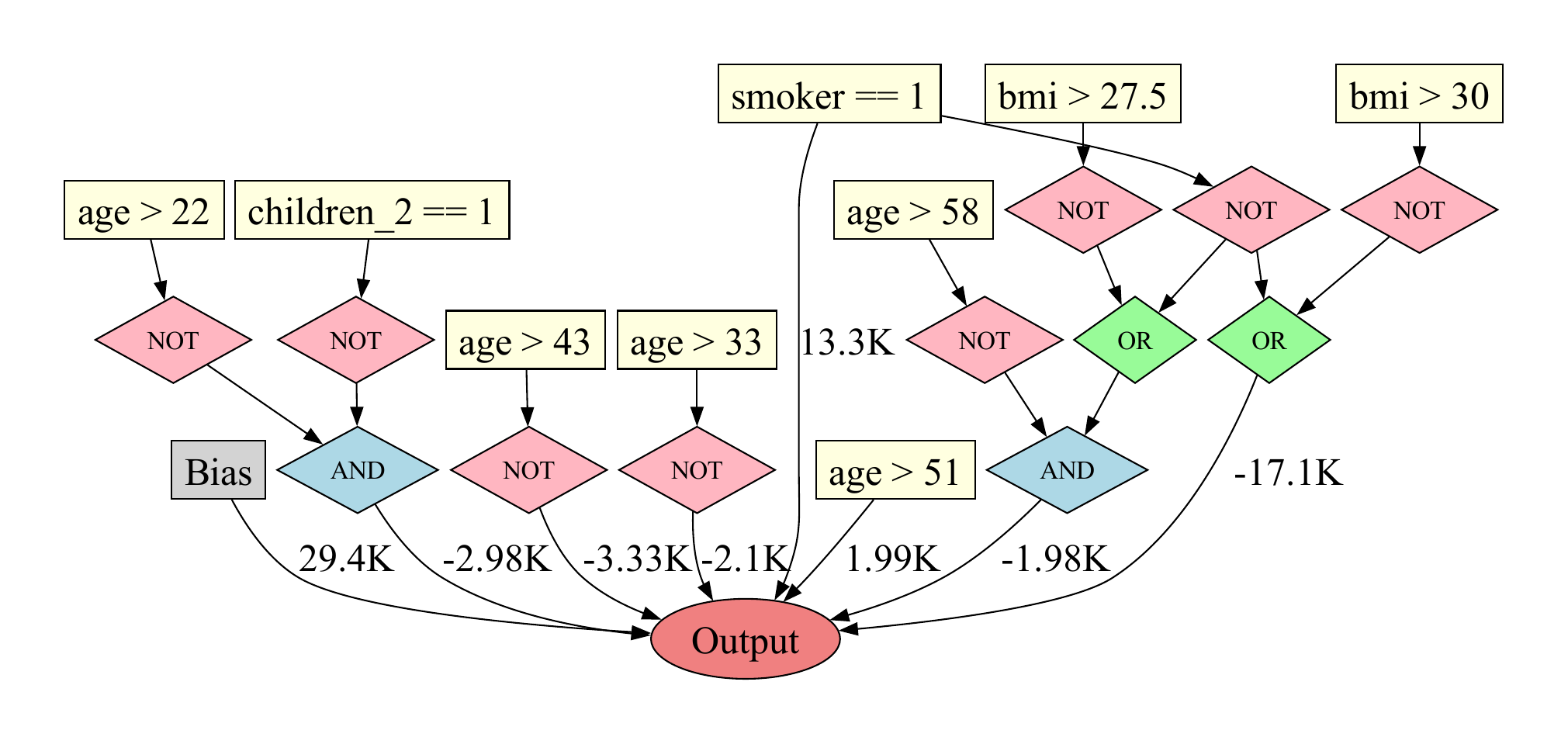}
\caption{Decision process learned by a DLN on the \textit{Insurance} dataset.  The model attains a test $R^{2}$ of
0.866 and relies on two continuous features and two one-hot encoded categorical features, selected from an original set
of two continuous and 12 one-hot features.}
\label{viz-Insurance}
\Description{A visualization of the decision process learned by a DLN on the Insurance dataset. 
The model uses two continuous features and two one-hot encoded categorical features, 
which were selected from a larger set, to achieve a test R-squared score of 0.866.}
\end{figure}

\begin{figure}[!htbp]
\centering
\includegraphics[width=\textwidth]{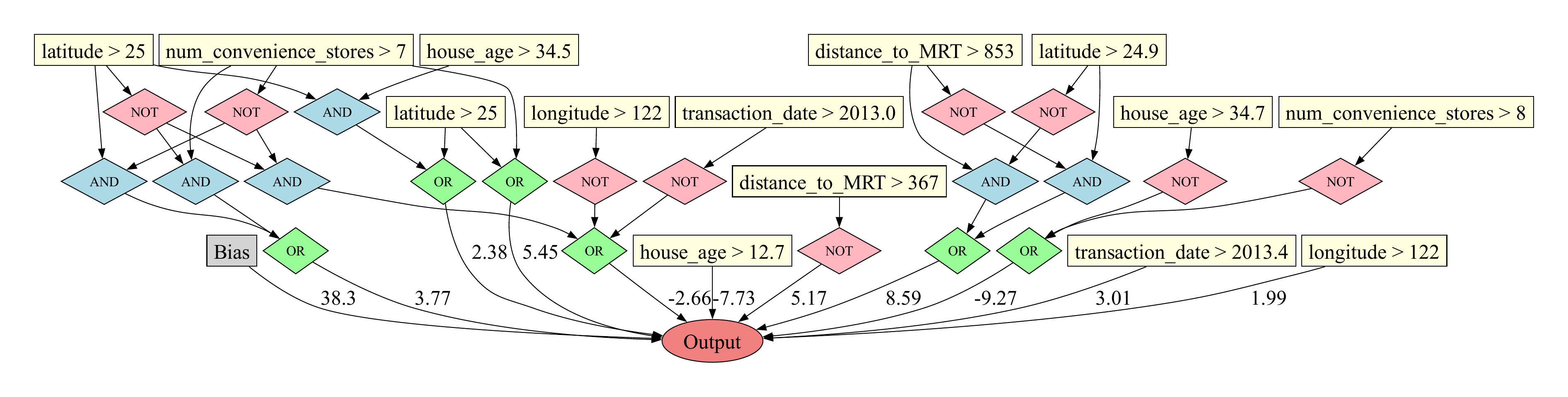}
\caption{Decision process learned by a DLN on the \textit{Estate} dataset.  The model attains a test $R^{2}$ of
0.758 and utilizes all six continuous input features.}
\label{viz-Estate}
\Description{A visualization of the logic learned by a DLN model for the Estate dataset. 
The diagram shows how the model uses six input features to make its final prediction.}
\end{figure}

\begin{figure}[!htbp]
\centering
\includegraphics[width=\textwidth]{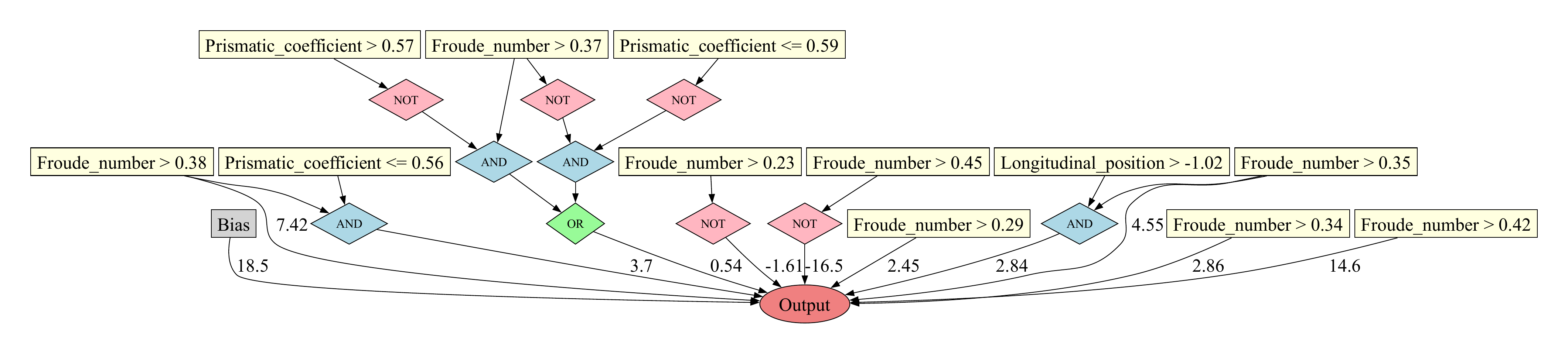}
\caption{Decision process learned by a DLN on the \textit{Yacht} dataset.  The model attains a test $R^{2}$ of 0.997
and selects three of the six available continuous features.}
\label{viz-Yacht}
\Description{A visualization of the decision process learned by a DLN on the Yacht dataset. 
The model selects three of the six available continuous features to achieve 
a test R-squared score of 0.997.}
\end{figure}

A trained regression DLN can be parsed directly and its prediction expressed as a weighted collection of logic
rules.  Figs.~\ref{viz-Insurance}, \ref{viz-Estate}, and \ref{viz-Yacht} illustrate three examples.  Yellow
rectangles represent input features with their learned thresholds; each outputs a binary value.  Diamonds contain
binary logic operators.  The network's final prediction is a weighted sum of
binary rule outputs, where each rule's coefficient is shown to the right of the edge that connects the rule node to the
output node.  Like their classification counterparts, regression DLNs perform implicit feature selection: In
Figs.~\ref{viz-Insurance} and \ref{viz-Yacht}, only a subset of the available features is used to achieve high accuracy.

\subsection{Ablation Studies}\label{exp-ablation}
We conduct ablation studies on four design choices: temperature scheduling, unified training phases, search-space
subsetting, and concatenation of ThresholdLayers to hidden LogicLayers.  Table~\ref{results-ablation} reports
mean results over 10 random seeds, each evaluated with 32 hyperparameter-search trials.  The configuration used for
the main experiments is labeled \textit{Orig}.  For context, the bottom rows of
Table~\ref{results-ablation} list the $R^{2}$ and OP ranks that each ablated variant would have achieved in
Tables~\ref{results-r2} and~\ref{results-ops}. \\
\textbf{Temperature scheduling.}~Tuning the initial temperature $\tau$ and its decay factor $\gamma$ is essential;
disabling this schedule markedly reduces performance. \\
\textbf{Training phases.}~While the original classification DLN benefited from two distinct phases, we find that a
single unified phase yields slightly better accuracy for regression DLNs. \\
\textbf{Search-space subset.}~Our default setting restricts each neuron to eight logic gates and eight input links, mirroring
the original DLN.  Expanding to 16 gates and 16 links offers comparable accuracy, whereas shrinking to four of each
degrades results. \\
\textbf{Input concatenation.}~Omitting the direct concatenation of ThresholdLayer outputs to hidden LogicLayers leads
to a substantial drop in accuracy.

\begin{table}[!htbp]
\centering
\caption{Average test $R^{2}$ for the ablation studies across 10 random seeds and 32 hyperparameter trials}
\label{results-ablation}
\small{
\begin{tabularx}{\textwidth}{l*{6}{Y}}
\toprule
 & Orig & No $\tau$ sched. & Two phases & Subspace 16 & Subspace 4 & No concat \\
\midrule
Abalone & \meanstd{0.525}{0.024} & \meanstd{0.524}{0.024} & \meanstd{0.526}{0.027} & \meanstd{0.522}{0.025} &
\meanstd{0.520}{0.024} & \meanstd{0.514}{0.015} \\
\addlinespace[2pt]
Airfoil & \meanstd{0.871}{0.021} & \meanstd{0.830}{0.024} & \meanstd{0.864}{0.016} & \meanstd{0.874}{0.014} & \meanstd{0.866}{0.015} & \meanstd{0.859}{0.018} \\
\addlinespace[2pt]
Bike & \meanstd{0.902}{0.016} & \meanstd{0.856}{0.037} & \meanstd{0.900}{0.017} & \meanstd{0.911}{0.016} & \meanstd{0.880}{0.017} & \meanstd{0.854}{0.039} \\
\addlinespace[2pt]
CCPP & \meanstd{0.942}{2.5e-03} & \meanstd{0.939}{2.8e-03} & \meanstd{0.941}{3.1e-03} & \meanstd{0.941}{3.0e-03} &
\meanstd{0.940}{2.9e-03} & \meanstd{0.938}{2.9e-03} \\
\addlinespace[2pt]
Concrete & \meanstd{0.887}{0.011} & \meanstd{0.866}{0.023} & \meanstd{0.876}{0.016} & \meanstd{0.882}{0.012} & \meanstd{0.885}{0.016} & \meanstd{0.873}{0.028} \\
\addlinespace[2pt]
Electrical & \meanstd{0.926}{7.2e-03} & \meanstd{0.913}{0.015} & \meanstd{0.932}{5.1e-03} & \meanstd{0.930}{8.6e-03} &
\meanstd{0.919}{8.0e-03} & \meanstd{0.918}{5.3e-03} \\
\addlinespace[2pt]
Energy & \meanstd{0.997}{4.3e-04} & \meanstd{0.998}{2.1e-04} & \meanstd{0.997}{4.4e-04} & \meanstd{0.997}{4.8e-04} & \meanstd{0.997}{4.2e-04} & \meanstd{0.998}{5.7e-04} \\
\addlinespace[2pt]
Estate & \meanstd{0.670}{0.060} & \meanstd{0.694}{0.057} & \meanstd{0.669}{0.050} & \meanstd{0.678}{0.063} & \meanstd{0.691}{0.059} & \meanstd{0.681}{0.073} \\
\addlinespace[2pt]
Housing & \meanstd{0.757}{0.017} & \meanstd{0.721}{0.014} & \meanstd{0.754}{0.011} & \meanstd{0.756}{0.019} &
\meanstd{0.751}{0.018} & \meanstd{0.732}{0.020} \\
\addlinespace[2pt]
Infrared & \meanstd{0.719}{0.051} & \meanstd{0.723}{0.046} & \meanstd{0.722}{0.027} & \meanstd{0.716}{0.044} & \meanstd{0.709}{0.045} & \meanstd{0.701}{0.043} \\
\addlinespace[2pt]
Insurance & \meanstd{0.854}{0.013} & \meanstd{0.849}{0.015} & \meanstd{0.850}{0.017} & \meanstd{0.853}{0.017} &
\meanstd{0.852}{0.015} & \meanstd{0.850}{0.016} \\
\addlinespace[2pt]
MPG & \meanstd{0.856}{0.016} & \meanstd{0.850}{0.022} & \meanstd{0.855}{0.029} & \meanstd{0.854}{0.023} & \meanstd{0.843}{0.025} & \meanstd{0.842}{0.029} \\
\addlinespace[2pt]
Parkinson's & \meanstd{0.798}{0.036} & \meanstd{0.690}{0.039} & \meanstd{0.783}{0.024} & \meanstd{0.819}{0.040} &
\meanstd{0.760}{0.047} & \meanstd{0.649}{0.057} \\
\addlinespace[2pt]
Wine & \meanstd{0.322}{0.017} & \meanstd{0.330}{0.016} & \meanstd{0.324}{0.023} & \meanstd{0.324}{0.019} & \meanstd{0.323}{0.019} & \meanstd{0.326}{0.023} \\
\addlinespace[2pt]
Yacht & \meanstd{0.997}{1.3e-03} & \meanstd{0.988}{9.9e-03} & \meanstd{0.996}{1.2e-03} & \meanstd{0.995}{3.7e-03} & \meanstd{0.996}{8.3e-04} & \meanstd{0.996}{1.7e-03} \\
\midrule
Mean $R^2$ $\uparrow$ & 0.802 & 0.785 & 0.799 & 0.803 & 0.795 & 0.782 \\
\addlinespace[1pt]
Average $R^2$ Rank $\downarrow$ & 3.53 & 4.07 & 3.60 & 3.47 & 3.60 & 3.87 \\
\midrule
Geo. Mean OPs $\downarrow$ & 29.9K & 26.7K & 27.5K & 27.6K & 27.5K & 18.3K \\
\addlinespace[1pt]
Average OPs Rank $\downarrow$ & 4.80 & 4.67 & 4.67 & 4.80 & 4.80 & 4.47 \\
\bottomrule
\end{tabularx}
}
\end{table}

\subsection{Limitations}
The main limitation of DLNs is slow training. This stems from the search for neuron functions and connections as a 
large differentiable classification problem. While the forward pass is efficient, this novel training procedure 
currently lacks built-in framework acceleration (e.g., in PyTorch). A second limitation is the lack of a full 
hardware validation on specialized accelerators. While we frame DLNs as promising candidates for hardware, this work is limited 
to demonstrating their efficiency on a CPU. Finally, a single DLN's accuracy can lag behind top-performing 
black-box models like XGBoost on several datasets, indicating a potential capacity limit, though we show this 
can be partially mitigated through ensembling.

\section{Conclusion and Future Directions}\label{sec-conclusion}
We have extended DLNs to tabular regression and shown that they preserve accuracy, interpretability,
and inference efficiency in this setting.  Two key training refinements, temperature scheduling and a unified
single-phase optimization, further improve performance. Our experiments also show that DLNs respond
well to simple ensembling, with a two-model ensemble outperforming random forest on both accuracy
and inference cost. This suggests that DLNs are promising candidates for integration into more
sophisticated ensemble frameworks. Promising avenues for future research include integrating
neural components into logic-based networks, as exemplified by convolutional LGN, 
which merges convolution and pooling operations with logical structures.  Another direction is to adapt DLNs to
additional data modalities, such as time series, thereby broadening their applicability.

\appendix
\section{Supplementary Results}\label{appendix-rmse}\label{appendix-mae}
We plot the RMSE and MAE metrics corresponding to Table~\ref{results-r2} in Tables~\ref{results-rmse} and \ref{results-mae}, 
respectively.  Across datasets, both RMSE and MAE are strongly and inversely correlated with
$R^{2}$.  DLN attains the third-best overall mean under both error measures.

\begin{table}[!htbp]
\centering
\caption{Average test RMSE across 10 random seeds and 128 hyperparameter trials per model}
\label{results-rmse}
\footnotesize{
\begingroup
\setlength{\tabcolsep}{1.5pt}
\begin{tabularx}{\textwidth}{l*{11}{Y}}
\toprule
& Linear & Ridge & Lasso & KNN & DT & AB & RF & XGB & SVR & MLP & DLN \\
\midrule
Abalone & 2.16 & 2.16 & 2.16 & 2.20 & 2.29 & 2.32 & 2.13 & 2.11 & 2.12 & 2.07 & 2.18 \\
\addlinespace[2pt]
Airfoil & 4.77 & 4.78 & 4.78 & 2.42 & 3.10 & 3.48 & 2.49 & 1.50 & 2.89 & 1.64 & 2.28 \\
\addlinespace[2pt]
Bike & 140 & 140 & 140 & 106 & 55.9 & 105 & 65.3 & 37.9 & 108 & 44.6 & 52.5 \\
\addlinespace[2pt]
CCPP & 4.54 & 4.54 & 4.54 & 3.52 & 4.11 & 4.80 & 3.90 & 3.11 & 3.94 & 3.86 & 4.06 \\
\addlinespace[2pt]
Concrete & 10.3 & 10.3 & 10.3 & 8.70 & 7.31 & 7.38 & 5.65 & 4.23 & 5.84 & 5.53 & 5.40 \\
\addlinespace[2pt]
Electrical & 0.0220 & 0.0220 & 0.0220 & 0.0166 & 0.0183 & 0.0154 & 0.0147 & 7.86e-03 & 7.19e-03 & 6.57e-03 & 9.36e-03 \\
\addlinespace[2pt]
Energy & 3.00 & 3.01 & 3.01 & 2.12 & 0.676 & 1.78 & 0.639 & 0.415 & 0.953 & 0.513 & 0.469 \\
\addlinespace[2pt]
Estate & 7.97 & 7.97 & 7.97 & 6.91 & 7.10 & 6.82 & 6.25 & 6.25 & 6.75 & 7.75 & 7.25 \\
\addlinespace[2pt]
Housing & 67.1K & 67.1K & 67.1K & 58.2K & 58.8K & 73.8K & 58.0K & 44.8K & 54.3K & 52.0K & 55.8K \\
\addlinespace[2pt]
Infrared & 0.256 & 0.254 & 0.253 & 0.276 & 0.257 & 0.243 & 0.239 & 0.238 & 0.251 & 0.287 & 0.261 \\
\addlinespace[2pt]
Insurance & 6.07K & 6.06K & 6.06K & 6.51K & 4.61K & 4.61K & 4.46K & 4.40K & 5.00K & 5.21K & 4.52K \\
\addlinespace[2pt]
MPG & 3.49 & 3.49 & 3.51 & 2.87 & 3.35 & 3.18 & 2.88 & 2.85 & 2.68 & 3.05 & 2.98 \\
\addlinespace[2pt]
Parkinsons & 9.67 & 9.67 & 9.67 & 6.23 & 2.99 & 7.86 & 3.15 & 1.69 & 6.65 & 4.28 & 3.99 \\
\addlinespace[2pt]
Wine & 0.718 & 0.718 & 0.718 & 0.693 & 0.744 & 0.723 & 0.691 & 0.678 & 0.683 & 0.729 & 0.713 \\
\addlinespace[2pt]
Yacht & 8.81 & 8.78 & 8.65 & 7.86 & 1.20 & 1.47 & 1.06 & 0.634 & 1.32 & 1.13 & 0.828 \\
\midrule
GMean RMSE $\downarrow$    & 9.84 & 9.83 & 9.83 & 8.10 & 6.13 & 7.46 & 5.64 & 4.30 & 6.20 & 5.29 & 5.35 \\ \addlinespace[1pt]
Average Rank $\downarrow$  & 9.07 & 9.20 & 8.93 & 6.53 & 6.53 & 7.33 & 3.67 & 1.27 & 4.47 & 4.53 & 4.47 \\
\bottomrule
\end{tabularx}
\endgroup
}
\end{table}

\begin{table}[!htbp]
\centering
\caption{Average test MAE across 10 random seeds and 128 hyperparameter trials per model}
\label{results-mae}
\begingroup
\setlength{\tabcolsep}{1.5pt}
\footnotesize{
\begin{tabularx}{\textwidth}{l*{11}{Y}}
\toprule
& Linear & Ridge & Lasso & KNN & DT & AB & RF & XGB & SVR & MLP & DLN \\
\midrule
Abalone & 1.57 & 1.57 & 1.57 & 1.54 & 1.62 & 1.76 & 1.50 & 1.50 & 1.47 & 1.48 & 1.57 \\
\addlinespace[2pt]
Airfoil & 3.69 & 3.69 & 3.69 & 1.63 & 2.34 & 2.83 & 1.91 & 0.993 & 2.05 & 1.16 & 1.71 \\
\addlinespace[2pt]
Bike & 105 & 105 & 104 & 69.1 & 33.4 & 82.9 & 41.9 & 22.9 & 67.3 & 27.5 & 36.5 \\
\addlinespace[2pt]
CCPP & 3.61 & 3.61 & 3.61 & 2.48 & 3.04 & 3.76 & 2.95 & 2.19 & 2.95 & 2.87 & 3.09 \\
\addlinespace[2pt]
Concrete & 8.23 & 8.25 & 8.27 & 6.43 & 5.22 & 6.01 & 4.27 & 2.86 & 4.29 & 3.83 & 3.98 \\
\addlinespace[2pt]
Electrical & 0.0174 & 0.0174 & 0.0174 & 0.0129 & 0.0142 & 0.0126 & 0.0116 & 5.74e-03 & 4.52e-03 & 4.38e-03 & 7.14e-03 \\
\addlinespace[2pt]
Energy & 2.13 & 2.13 & 2.14 & 1.39 & 0.438 & 1.47 & 0.432 & 0.292 & 0.722 & 0.360 & 0.347 \\
\addlinespace[2pt]
Estate & 6.07 & 6.08 & 6.06 & 4.88 & 5.16 & 4.99 & 4.43 & 4.35 & 4.73 & 5.51 & 5.32 \\
\addlinespace[2pt]
Housing & 48.9K & 48.9K & 48.9K & 39.0K & 39.3K & 55.8K & 40.2K & 29.2K & 35.9K & 35.0K & 39.1K \\
\addlinespace[2pt]
Infrared & 0.203 & 0.200 & 0.199 & 0.207 & 0.198 & 0.189 & 0.185 & 0.185 & 0.192 & 0.221 & 0.201 \\
\addlinespace[2pt]
Insurance & 4.24K & 4.24K & 4.21K & 3.80K & 2.15K & 2.93K & 2.52K & 2.49K & 2.93K & 3.12K & 2.54K \\
\addlinespace[2pt]
MPG & 2.67 & 2.67 & 2.66 & 2.05 & 2.41 & 2.33 & 2.01 & 2.01 & 1.89 & 2.10 & 2.15 \\
\addlinespace[2pt]
Parkinsons & 8.03 & 8.03 & 8.04 & 3.99 & 1.41 & 6.89 & 2.10 & 0.849 & 4.78 & 2.54 & 2.99 \\
\addlinespace[2pt]
Wine & 0.555 & 0.556 & 0.556 & 0.538 & 0.578 & 0.571 & 0.538 & 0.525 & 0.524 & 0.566 & 0.552 \\
\addlinespace[2pt]
Yacht & 7.18 & 7.10 & 6.87 & 4.11 & 0.644 & 1.02 & 0.601 & 0.381 & 0.950 & 0.611 & 0.523 \\
\midrule
GMean MAE $\downarrow$    & 7.54 & 7.53 & 7.51 & 5.52 & 4.06 & 5.75 & 3.94 & 2.86 & 4.34 & 3.58 & 3.84 \\ \addlinespace[1pt]
Average Rank $\downarrow$ & 9.40 & 9.33 & 9.13 & 5.80 & 5.93 & 7.73 & 3.73 & 1.53 & 4.07 & 4.33 & 5.00 \\
\bottomrule
\end{tabularx}
}
\endgroup
\end{table}

\clearpage
\section*{Acknowledgments}
This work was supported by NSF under Grant No. CCF-2416541.

\bibliographystyle{unsrtnat}
\bibliography{references}

\end{document}